\documentclass[12pt]{article}

\usepackage[english]{babel}
\usepackage[utf8x]{inputenc}
\usepackage[T1]{fontenc}

\usepackage[a4paper,top=3cm,bottom=2cm,left=3cm,right=3cm,marginparwidth=1.75cm]{geometry}
\usepackage{amsmath}
\usepackage{graphicx}
\usepackage[colorinlistoftodos]{todonotes}
\usepackage[colorlinks=true, allcolors=blue]{hyperref}

\usepackage{multirow}

\title{Learning and Acting in Peripersonal Space: \\  Moving, Reaching, and Grasping\thanks{This work has taken place in the Intelligent Robotics Lab in the Computer Science and Engineering Division of the University of Michigan.  Research of the Intelligent Robotics lab is supported in part by a grant from the National Science Foundation (IIS-1421168).}}
\author{Jonathan Juett and Benjamin Kuipers\thanks{\{jonjuett,kuipers\}@umich.edu, Computer Science \& Engineering, University of Michigan, Ann Arbor MI 48109 USA}}

\begin{document}
\maketitle

\begin{abstract}
The young infant explores its body, its sensorimotor system, and the immediately accessible parts of its environment, over the course of a few months creating a model of peripersonal space useful for reaching and grasping objects around it.  Drawing on constraints from the empirical literature on infant behavior, we present a preliminary computational model of this learning process, implemented and evaluated on a physical robot.  

The learning agent explores the relationship between the configuration space of the arm, sensing joint angles through proprioception, and its visual perceptions of the hand and grippers.  The resulting knowledge is represented as the peripersonal space (PPS) graph, where nodes represent states of the arm, edges represent safe movements, and paths represent safe trajectories from one pose to another.

In our model, the learning process is driven by intrinsic motivation.  When repeatedly performing an action, the agent learns the typical result, but also detects unusual outcomes, and is motivated to learn how to make those unusual results reliable.  Arm motions typically leave the static background unchanged, but occasionally bump an object, changing its static position.  The reach action is learned as a reliable way to bump and move a specified object in the environment.

Similarly, once a reliable reach action is learned, it typically makes a quasi-static change in the environment, moving an object from one static position to another.  The unusual outcome is that the object is accidentally grasped (thanks to the innate Palmar reflex), and thereafter moves dynamically with the hand.  Learning to make grasping reliable is more complex than for reaching, but we demonstrate significant progress.

Our current results are steps toward autonomous sensorimotor learning of motion, reaching, and grasping in peripersonal space, based on unguided exploration and intrinsic motivation.

\end{abstract}

\section{Introduction}

The newborn human baby moves its limbs, apparently aimlessly.  After a few months of unguided exploration, it can reliably {\em reach} (moving a hand to a visually perceived object) and {\em grasp} (taking control of the object and moving it with the hand).  The young child gains skill at reaching and grasping, learning to move objects dextrously from one location to another as desired.  This is a major step in achieving mastery over its environment.

{\em Peripersonal space (PPS)} refers to the space immediately around an agent, accessible to its arms and hands for the manipulation of objects.  Robots are typically programmed by expert humans to access peripersonal space, exploiting computer vision algorithms to identify the locations of objects, and forward and inverse kinematics algorithms to plan motion of the arms.  These tools are not available to young infants, who nonetheless achieve a level of robust mastery over peripersonal space that robots cannot match.

\begin{figure}
\hrule
\vspace{1mm}
\begin{center}
\includegraphics[width=\textwidth]{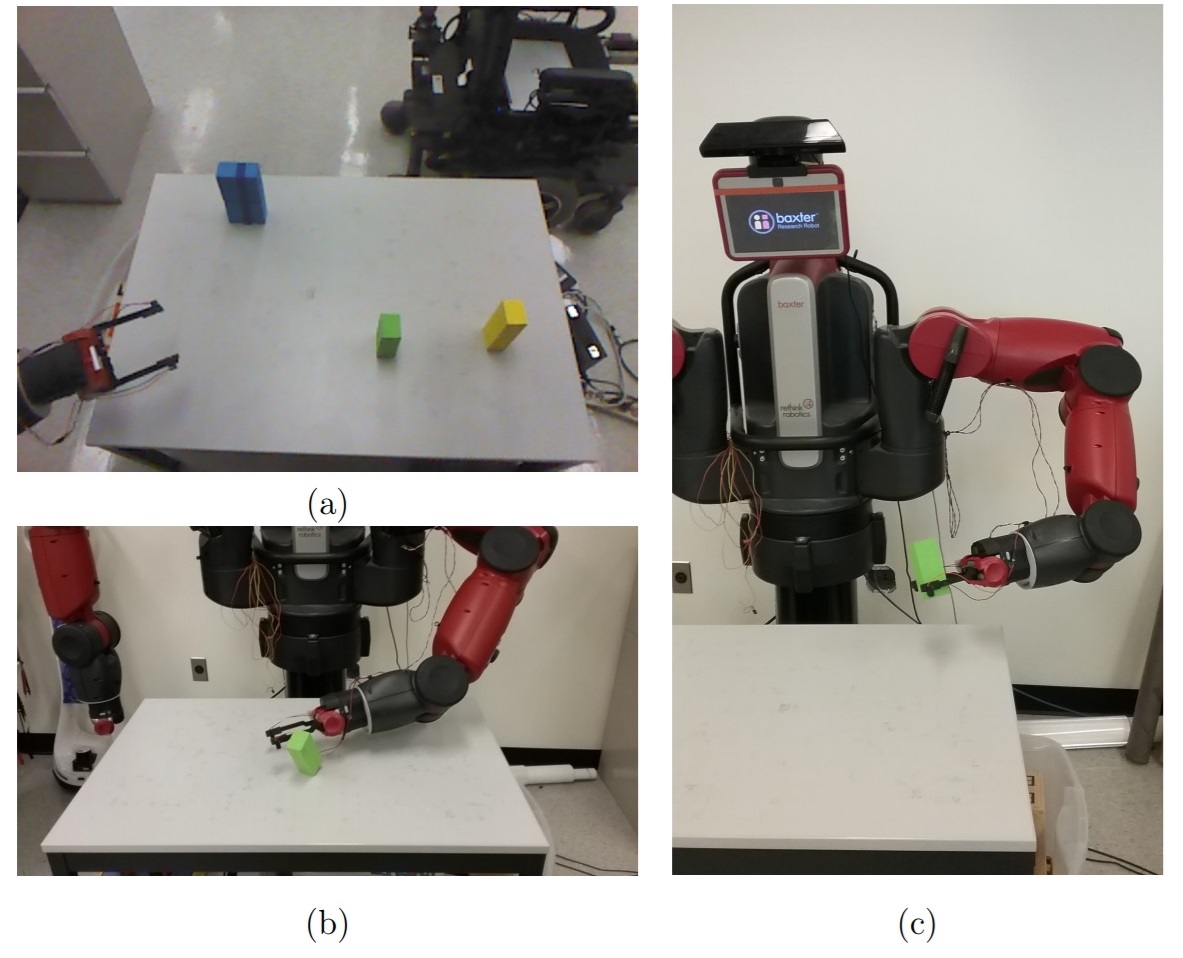}
\end{center}
\vspace{-4mm}
\caption{\textbf{(a)} The environment from the agent's perspective, including simple block objects used in this work. \textbf{(b)} The unusual \textit{bump} event is observed during random motions when the hand collides with an object and changes its state. 
The agent defines a $reach$ action, which uses trajectories that share properties with the accidental bumps to make purposeful repetition of the rare event reliable. Actions are evaluated with a single foreground object present. \textbf{(c)} Our model is implemented on, and evaluated with, a physical Baxter robot. The \textit{grasp} action allows the agent to continue to control the motion of an object as it travels with the hand. The first grasps occur accidentally during reaching, and incorporation of additional features makes intentional grasps moderately reliable.}
\label{fig:intro}
\vspace{1mm}
\hrule
\end{figure}

In this paper, we present a preliminary computational model, implemented and evaluated on a physical robot (Figure~\ref{fig:intro}), describing in detail how an embodied agent might explore the capabilities of its body, its sensorimotor system, and the structure of its peripersonal space to learn to reach and grasp.

Our current preliminary model is a proof of concept.  Our successful robot implementation demonstrates the ability of an agent to achieve reasonable levels of reliability at reaching and grasping from unguided experience, through a learning process driven by intrinsic motivation.  While our model is consistent with important properties of human infant learning, other observed properties have not yet been incorporated into our model, though we plan to do so in future work.

\subsection{Overview}

The agent’s problem is to learn a useful representation for knowledge of peripersonal space that it can acquire from its own experience doing unguided exploration.  For peripersonal space, a useful representation necessarily integrates several high-dimensional concrete and abstract spaces, including the two-dimensional retinal images that define visual space, the vector of proprioceptive joint-angle sensors in the arm, the three-dimensional workspace around the agent, and the configuration space of the arm (six or more dimensions).  This includes and expands upon the problem of ``multisensory integration'' as discussed in the psychological literature (e.g., \cite{Bremner-tics-08,Corbetta-fp-14}).  At the same time, our current model assumes that the posture of the robot's body, and the {\em pose} (position and orientation) of its camera(s) are fixed.  The robot's only motor action is to control the joints of its arm.

Taking the perspective of developmental robotics, we assume that our learning agent has access to raw sensory input (proprioceptive joint angles and raw visual images), and can send incremental motor commands to its joints, but that it has no prior geometric model of its body, its workspace, or the configuration space of its arms.  

The robot begins exploring the configuration space of the arm with apparently aimless ``motor babbling.''  For human infants, it is known that these motions are not completely random, but are biased toward keeping the hand in view \cite{vonHofsten-dp-82,vonHofsten-dp-84,vanderMeer-science-95,vanderMeer-ejpn-97}.  Our model starts with this constrained random exploration, building descriptions of experienced states in terms of both the vector of joint angles and the visual observation.

Self-generated intermediate goals in the learning process will include the abilities to {\em reach} (bumping and changing the position of a perceived object) and to {\em grasp} (taking control of an object and being able to move it with the hand).  

As the robot explores its configuration space, it creates a graph that we call the {\em Peri-Personal Space (PPS) graph}.  A node $n_i$ is created for the PPS graph for each state of the arm after each random motion.  The node $n_i$ is defined by the vector $q_i$ of joint angles at that state, so the node represents a point in the configuration space of the arm.  The node is also associated with a visual percept $P(n_i)$, which at this time is the pair of 2D visual images the robot receives from its RGB-D sensor:  one 2D RGB camera image $P_{RGB}$, and a second 2D image $P_D$ (with the same frame of reference) representing the sensed distances at each point in the first image.  Edges in the PPS graph represent the opportunity for safe motion between two nodes.  Safe motion from one node to the next implies the existence of an edge.  A new node can also be connected to existing nodes whose joint angle vectors are close enough, under the assumption that a short move will be safe.  By construction, a path in the PPS graph represents a safe potential trajectory from an initial to a final node.  In our current experiments, the robot constructs a PPS Graph of about 3000 nodes to describe its peripersonal space.

In the current model, the robot follows an edge from one node to another by linear interpolation from the joint angle vector of one node, to the joint angle vector of the next.  In future work, we plan to model human infant motor control more realistically by implementing such motion as a dynamical control law \cite{Thelen-cd-93}.

While exploring its configuration space, the robot observes the typical effects of its actions on the visually-sensed scene.  Most often, the arm and hand move against the static background, while everything else remains fixed.  However, in unusual cases, the hand bumps and moves an object to a new static pose.  Our model includes a kind of {\em intrinsic motivation}:  upon observing an unusual event, the learning agent searches for a way to accomplish that event reliably.  If successful, it transforms an unusual event into a potentially useful {\em action}.  In our model, this pattern is applied twice:  once to learn {\em reaching}, and once to learn {\em grasping}.

For learning to reach and bump a block in the robot’s environment, the unusual event is a quasi-static change in the block’s observed position, observed when comparing visual images collected before and after a hand (and arm) motion of the robot.   To make this reliable, given an observed block in the environment, the robot needs a criterion for selecting a target node in the PPS graph, such that moving to that node will reliably reach and bump the observed block.  As we shall see, the robot can define a suitable search space, and identify a criterion for selecting an appropriate target node, increasing the reliability of the Reach action.

Each node $n$ in the PPS graph $G$ represents a configuration of the arm, and includes an image of the hand in the otherwise-empty static environment.  After $G$ has been created, a situation $s$ includes a visual image of a block added to the static environment.  A feature-comparison function $f(n,s)$ compares the binary images of the hand (in $n$) and the block (in $s$), and measures the degree of match.  We can then search over $G$ to find the target node that maximizes $f(n,s)$.
\begin{equation}
n_f^*(s) = \arg \max_{n\in G} f(n,s)
\end{equation}

To evaluate a feature-comparison function $f$ over a set $S$ of situations, we determine the probability that moving to the selected target node $n_f^*(s)$ results in $Bump(s)$, bumping and moving the block.

The learning process searches over a set $F$ of possible feature-comparison functions appropriate to this task.
\begin{equation}
 f^* = \arg \max_{f\in F} P(Bump(s) \; | \; s\in S \wedge MoveTo(n^*_f(s))
\end{equation}

In our first paper \cite{Juett-humanoids-16}, the visual system consisted of three independently-located monocular cameras, and $f^*$ maximized the IOU (intersection over union) measures of hand and block across all three images.  In our second paper \cite{Juett-iros-18}, the visual system consisted of a single RGB-D camera, and $f^*$ maximized the overlap of the D images, restricted to the intersection of the RGB images of hand and object.  In this paper, $f^*$ selects the node with the smallest center-to-center distance between hand and object, among those with nonempty intersections in both RGB and D images.

Since the configuration space has six or more dimensions, a PPS graph of several thousand nodes is still a relatively sparse approximation to the set of arm configurations.  However, by using the information on the neighbors of a given PPS node, we can approximate the {\em local Jacobian} matrix at that node, predicting how incremental changes to the joint angles will perturb the visual image of the hand.  The local Jacobian thus helps ``fill the gaps'' between the nodes of the PPS Graph, improving the agent's representation of its continuous, high-dimensional configuration space.  By adjusting the motion along the final edge in a trajectory to a target node, the reliability of a Reach action approaches $100\%$.

With a sequence of random actions, an accidental grasp would be extremely unlikely, requiring two specific random motions in sequence:  one to put the hand in the right place (say, probability $\epsilon$), and the second one immediately after, to close the fingers around the object (probability $\epsilon$).  The probability of the sequence of two actions is $\epsilon^2$, which is very small.  Fortunately, very young children have the {\em Palmar reflex} \cite{Futagi-ijp-12}, which responds to a touch to the palm by closing the fingers, wrapping them tightly around whatever touched the palm.  The Palmar reflex is present at birth, and fades away by about five months of age.  While it is present, the probability of an ``accidental grasp’’ is $\epsilon$, rather than $\epsilon^2$, which makes the unusual grasping event likely enough for learning.  We implement a simulated Palmar reflex on our Baxter robot with a break-beam sensor attached to the gripper fingers.

Once the robot learns a reliable Reach action, it continues to explore its environment by reaching to bump nearby blocks.  The unusual event is to trigger the Palmar reflex and thus to spontaneously grasp the block.  This is detected because the block, rather than being displaced from its static position to a new static position, becomes temporarily bound to the hand, and moves along with the hand.

Reliable grasping is much more difficult than reliable reaching, even exploiting the information in the PPS Graph.  In the long run, it will involve identifying the shape of the target object, learning to identify useful grasp points on that shape, and pre-shaping the hand to grasp that type of object securely.  For our experiments, with simple blocks, our methods have so far allowed us to reach about 50\% reliable grasping.  

\section{Related Work}

\subsection{The human model:  evidence from child development}

There is a rich literature in developmental psychology on how infants learn to reach and grasp, in which the overall chronology of learning to reach is reasonably clear (e.g., \cite{Berthier-iccs-11,Corbetta-fp-14}).  From birth to about 15 weeks, infants can respond to visual targets with ``pre-reaching'' movements that are generally not successful at making contact with the targets.  From about 15 weeks to about 8 months, reaching movements become increasingly successful, but they are jerky with successive submovements, some of which may represent corrective submovements \cite{vonHofsten-jmb-91}, and some of which reflect underdamped oscillations on the way to an equilibrium point \cite{Thelen-cd-93}.  For decades, early reaching was generally believed to require visual perception of both the hand and the target object, with reaching taking place through a process of bringing the hand and object images together (``visual servoing'').  However, a landmark experiment \cite{Clifton-cd-93} showed that the pattern and success rate of reaching by young infants is unaffected when the hand is not visible.  Toward the end of the first year, vision of the hand becomes important for configuring and orienting the hand in anticipation of contact with target objects.  The smoothness of reaching continues to improve over early years, toward adult reaches which typically consist of ``a single motor command with inflight corrective movements as needed'' \cite{Berthier-iccs-11}.

Theorists grapple with the problem that reaching and grasping require learning useful mappings between visual space (two- or three-dimensional) and the configuration space of the arm (with dimensionality equal to the number degrees of freedom).

Bremner, Holmes \& Spence \cite{Bremner-tics-08} address this issue under the term, {\em multisensory integration}, focusing on sensory modalities including touch, proprioception, and vision.  They propose two distinct neural mechanisms.  The first assumes a fixed initial body posture and arm configuration, and represents the positions of objects within an egocentric frame of reference.  The second is capable of re-mapping spatial relations in light of changes in body posture and arm configuration, and thus effectively encodes object position in a world-centered frame of reference.

Corbetta, et al \cite{Corbetta-fp-14} focus directly on how the relation is learned between proprioception (``the feel of the arm'') and vision (``the sight of the object'') during reach learning.  They describe three theories:  vision first; proprioception first; and vision and proprioception together.  Their experimental results weakly supported the proprioception-first theory, but all three had strengths and weaknesses.

Thomas, Karl \& Whishaw \cite{Thomas-fp-15} closely observed spontaneous self-touching behavior in infants during their first six months.  Their analysis supports two separately-developing neural pathways, for Reach, which moves the hand to contact the target object, and for Grasp, which shapes the hand to gain successful control of the object.

These and other investigators provide valuable insights into distinctions that contribute to answering this important question.  But different distinctions from different investigators can leave us struggling to discern which differences are competing theories to be discriminated, and which are different but compatible aspects of a single more complex reality.  

We believe that a theory of a behavior of interest (in this case, learning from unguided experience to reach and grasp) can be subjected to an additional demanding evaluation by working to define and implement a computational model capable of exhibiting the desired behavior.  In addition to identifying important distinctions, this exercise ensures that the different parts of a complex theory can, in fact, work together to accomplish their goal.

The model we present at this point is preliminary.  To implement it on a particular robot, certain aspects of the perceptual and motor system models will be specific to the robot, and not realistic for a human infant.  To design, implement, debug, and improve a complex model, we focus on certain aspects of the model, while others remain over-simplified.  For example, our model of the Peri-Personal Space (PPS) Graph uses vision during the creation of the PPS Graph, but then does not need vision of the hand while reaching to a visible object \cite{Clifton-cd-93}.  The early reaching trajectory will be quite jerky because of the granularity of the edges in the PPS Graph \cite{vonHofsten-jmb-91}, but another component of the jerkiness could well be due to underdamped dynamical control of the hand as it moves along each edge \cite{Thelen-cd-93}, which is not yet incorporated into our model.

\subsection{Robot developmental learning to reach and grasp}

\paragraph{Robotic Modeling.}

Some robotics researchers (e.g., \cite{Hersch-ijhr-08,Sturm-icra-08}) focus on learning the kind of precise model of the robot that is used for traditional forward and inverse kinematics-based motion planning.  Hersch, Sauser \& Billard \cite{Hersch-ijhr-08} learn a body schema for a humanoid robot, modeled as a tree-structured hierarchy of frames of reference, assuming that the robot is given the topology of the network of joints and segments and that the robot can perceive and track the 3D position of each end-effector.  Sturm, Plagemann \& Burgard \cite{Sturm-icra-08} start with a pre-specified set of variables and a fully-connected Bayesian network model.  The learning process uses visual images of the arm while motor babbling, exploiting markers that allow extraction of 6D pose for each joint.  Bayesian inference eliminates unnecessary links and learns probability distributions over variable values.  A goal of our model is to make much weaker assumptions about the variables and constraints in the model, and the information available from visual perception.

\paragraph{Neural Modeling.}

Other researchers (e.g., \cite{Oztop-ebr-04,Savastano-tamd-13}) structure their models according to hypotheses about the neural control of reaching and grasping, with constraints represented by neural networks that are trained from experience.  Oztop, Bradley \& Arbib \cite{Oztop-ebr-04} draw on empirical data from the literature about human infants, to motivate their computational model (ILGM) of grasp learning.  The model consists of neural networks representing the probability distributions of joint angle velocities.  They evaluate the performance of their model with a simulated robot arm and hand, assuming that reaching is already programmed in.  Their model includes a Palmar reflex, and they focus on learning an open-loop controller that is likely to terminate with a successful grasp.  Savastano and Nolfi \cite{Savastano-tamd-13} describe an embodied computational model implemented as a recurrent neural network, and evaluated on a simulation of the iCub robot.  They demonstrate pre-reaching, gross-reaching, and fine-reaching phases of learning and behavior, qualitatively matching observations of children such as diminished use of vision in the first two phases, and proximal-then-distal use of the arm's degrees of freedom.  The transitions from one phase to the next are represented by manually adding certain links and changing certain parameters in the network, begging the question about how and why those changes take place.

\paragraph{Sensorimotor Learning.}

Several recent research results are closer to our approach, in the sense of focusing on sensorimotor learning without explicit skill programming, exploration guidance, or labeled training examples.  Each of these (including ours) makes simplifying assumptions to support progress at the current state of the art, but each contributes a ``piece of the puzzle'' for learning to reach and grasp.

Jamone, et al \cite{Jamone-ras-14} define a Reachable Space Map over gaze coordinates (head yaw and pitch, and eye vergence (to encode depth)) during fixation. The control system moves the head and eyes to place the target object at the center of both camera images.  In the Reachable Space Map, $R=0$ describes unreachable targets; intermediate values describe how close manipulator joints are to the physical limits of their ranges; and $R=1$ means that all joints are well away from their limits.  The Reachable Space Map is learned from goal-directed reaching experience trying to find optimal reaches to targets in gaze coordinates.  Intermediate values of $R$ can then be used as error values to drive other body-pose degrees of freedom (e.g., waist, legs) to improve the reachability of target objects.  Within our framework, the Reachable Space Map would be a valuable addition (in future work), but the PPS Graph \cite{Juett-humanoids-16} is learned at a developmentally earlier stage of knowledge, before goal-directed reaching has a meaningful chance of success.  The PPS Graph is learned during non-goal-directed motor babbling, as a sampled exploration of configuration space, accumulating associations between the joint angles determining the arm configuration and the visual image of the arm.

Ugur, et al \cite{Ugur-tamd-15} demonstrate autonomous learning of behavioral primitives and object affordances, leading up to imitation learning of complex actions.  However, they start with the assumption that peripersonal space can be modeled as a 3D Euclidean space, and that hand motions can be specified via starting, midpoint, and endpoint coordinates in that 3D space.  Our agent starts with only the raw proprioceptively sensed joint angles in the arm, and the 2D images provided by vision sensors, and the PPS graph represents a learned mapping between those spaces.  The egocentric Reachable Space Map \cite{Jamone-ras-14} could be a step toward a 3D model of peripersonal space.

M. Hoffmann, et al \cite{Hoffmann-icdler-17} integrate empirical data from infant experiments with computational modeling on the physical iCub robot.  Their model includes haptic and proprioceptive sensing, but not vision.  They model the processes by which infants learn to reach to different parts of their bodies, prompted by buzzers on the skin.  They report results from experiments with infants, and derive constraints on their computational model.  The model is implemented and evaluated on an iCub robot with artificial tactile-sensing skin.  However, the authors themselves describe their success as partial, observing that the empirical data, conceptual framework, and robotic modeling are quite disparate, and not well integrated.  They aspire to implement a version of the sensorimotor account, but they describe their actual model as much closer to traditional robot programming.

\paragraph{Deep Network Learning.}

Convolutional neural networks (CNNs) have been a very powerful technology for supervised learning for visual recognition.  Pinto and Gupta \cite{Pinto-icra-16} showed how to apply CNNs to {\em self}-supervised learning, where a physical robot attempts to grasp visually-perceived objects and generates its own feedback from the success or failure of the grasp attempt.  With a multi-stage learning process, and after spending 700 hours self-generating 50K training examples, their robot achieved state-of-the-art (2015) performance.  

Inspired by this, Levine and colleagues \cite{Levine-arxiv-16,Levine-ijrr-18} extended the method to guide the hand continuously using visual servoing, and trained it in parallel on up to 14 manipulator arms, scaling up to 800K training examples, and achieving state-of-the-art (2016) performance.  (The tasks and conditions for these two approaches are somewhat different.)

Deep CNN learning gives impressive results at selecting useful grasp points from visual information, but the structures of the reaching and grasping actions are programmed by humans.  It is notable that the robot learns to grasp by visual servoing, whereas an important result about early reaching in human infants is that grasping can succeed even without vision of the hand.  We don't yet understand why human reaching and grasping is learned the way it is, but it may well be for a good reason, considering that humans are still far more dextrous and skilled at grasping than even the best robots.

\section{Notation}

In order to describe our model in detail, we will need more detailed and specific notation, which is described in this section.

Our agent acts in the world with control over the left arm of a Baxter Research Robot. The state of this arm can be given by eight degrees of freedom, a set of seven joint angles, $q = \{q^1, ... q^7\}$ and the aperture between the gripper fingers, described by a percentage of its maximum width, $a$. The agent's representation of space relies on building a PPS Graph $G = (\{n\},\{e\})$ as described in the following section, where each node $n_i$ stores a particular $q_i$ and $a_i$ for a visited state. All of this information can be directly sensed through proprioception. Each edge $e_{i,i'}$ is denoted by the indices of its endpoint nodes.

The agent also has the ability to make and save visual percepts of its environment. Each visual percept $P$ is taken by a fixed-viewpoint RGB-D camera, providing an RGB image $P_{RGB}$ and a depth-registered image $P_D$. During the construction of $G$, the agent may save a percept $P\left(n_i\right)$ taken while it is paused at $n_i$. After $G$ is complete, the agent may plan a trajectory for the arm's motion as a sequence of nodes $T = \langle n_{T_1}, ... , n_{T_{\vert T \vert}}\rangle $ that forms a graph path in $G$. The agent may later choose to smooth or fine-tune this trajectory, but initially executes the planned trajectory by visiting each $n_{T_j}$ in order, that is, setting the joint angles and gripper aperture to the stored $q_{T_j}$ and $a_{T_j}$. The agent may take new visual percepts at each $n_{T_j}$, denoting them $P'\left(n_{T_j}\right)$ since $P\left(n_i\right)$ already exists for $i = T_j$. The original percept for the node often gives a clearer, unoccluded image of the hand so it is not replaced. The final and penultimate nodes of the trajectory will be denoted $n_f \equiv n_{T_{\vert T \vert}}$ and $n_p \equiv n_{T_{\vert T \vert - 1}}$, respectively. All trajectories begin at the agent's home node, denoted $n_h \equiv n_{T_1}$. For convenience, the subscripts $f$, $p$, and $h$ may be used to identify other features associated with the final, penultimate, and home nodes. All actions taken in this work involve a return trip along the chosen trajectory, visiting each node in $T$ in reverse order. Observations made during the return are denoted $P''\left(n_{T_j}\right)$.

The agent is capable of processing the visual percepts to create a more efficient representation of the significant objects in each image, corresponding to the self and foreground objects that may be targets of manipulation actions. For each $n_i \in G$, the agent finds the end effector in $P_{RGB}(n_i)$ and records two binary masks that describe its location in the image. The {\em palm mask} $p_i$ is defined to be the region between the gripper fingers, which will be most relevant for grasping.\footnote{We use the word ``palm'' for this region because of its functional (though not anatomical) similarity to the human palm, especially as the site of the Palmar reflex \cite{Futagi-ijp-12}.} The {\em hand mask} $h_i$ includes this region as well as the gripper fingers and the wrist near the base of the hand. $h_i$ reflects the full space occupied by the hand, and is better suited to identify nodes describing hand positions that may collide with obstacles. The state representation for a node also includes the range of depths the end effector is observed to occupy. This range is found by indexing into $P_{D}(n_i)$ with either mask, and determining the minimum and maximum depth values over these pixels. That is, the depth range of the palm $D(p_i) \equiv [\textnormal{min}(P_{D}(n_i)[p_i]), \textnormal{max}(P_{D}(n_i)[p_i])]$, and the depth range of the full hand $D(h_i)$ is defined analogously.

\begin{figure}
\hrule
\vspace{1mm}
\begin{center}
\includegraphics[width=\textwidth]{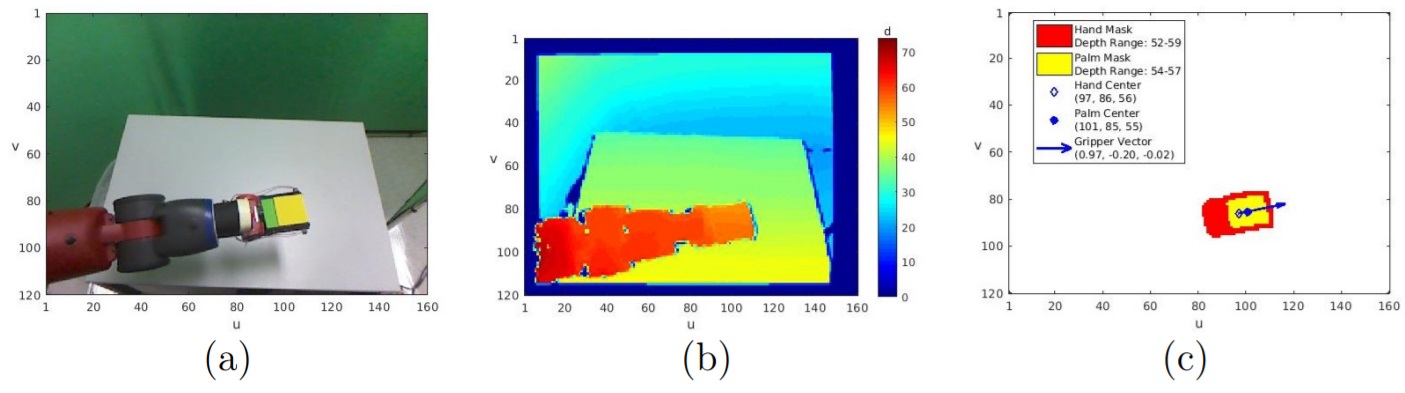}
\end{center}
\vspace{-4mm}
\caption{An example of the agent's visual percept and stored representation for a node. \textbf{(a)} A single RGB image, scaled down to 120x160 resolution, taken while the arm configuration is set to the joint angles stored for the node. \textbf{(b)} The registered depth image taken at the same time. Note that the depth values are a measure of disparity, so smaller values are further from the camera. \textbf{(c)} The full representation the agent will store for this node. Aided by the yellow block held between the gripper fingers, the agent segments the palm mask, corresponding to the grasping region of the hand. The larger hand mask includes the palm mask (shown in yellow) and parts of the robot image segment near the block, typically the gripper fingers and lower wrist (shown in red). The range of values that exist in the depth image within each mask is also stored, as are the center of mass and mean depth value for each mask. Finally, to estimate the direction the grippers are pointing, a vector is drawn from the hand mask center through the palm mask center.}
\label{fig:NodeRepresentation}
\vspace{1mm}
\hrule
\end{figure}

For some purposes, it will be easier for the agent to work with an abstracted center and direction of the end effector, and it can derive these from the binary masks and depth ranges. Centers and directions will have three components, two for the $(u,v)$-coordinates in the RGB image, and one $(d)$ for depth values in the Depth image. For a node $n_i$, the center of the palm $c^p_i$ is composed of the center of mass of $p_i$ and the average depth, $\textnormal{mean}(P_{D}(n_i)[p_i])$, and the center of the hand $c^h_i$ is derived from $h_i$ and $P_{D}(n_i)[p_i]$ in the same manner. To describe the direction the gripper is facing, the agent can use the relationship between the two centers. $h_i$ includes approximately the same amount of extra area on each side of the palm since the grippers are thin and appear with similar widths for most poses of the hand. However, $h_i$ includes a relatively large area at the base of the hand, and almost no pixels past the palm since the palm extends to the tip of the hand. As a result, a vector $\vec{g}_i$ drawn from $c^h_i$ through $c^p_i$ goes from the proximal to distal portion of the hand, and is near parallel to the gripper fingers. Figure~\ref{fig:NodeRepresentation} displays an example percept $P$ and these components derived from it to complete the representation of a node.

It is also necessary for the agent to be able to represent any nonself foreground objects. In this work, these objects are introduced after $G$ is completely constructed, and the initial observations $P$ have been taken, so they will appear only in observations of the forward trajectory $P'$ and observations of the reverse trajectory $P''$. As these objects will be targets of the agent's manipulation actions, $t$ will refer to the binary mask where an object is seen. When multiple objects are present, $t'_{k,j}$ denotes the $k$th object's mask in $P'_{RGB}(n_{T_j})$, and $t''_{k,j}$ denotes the $k$th object's mask in $P''_{RGB}(n_{T_j})$. If only one object is present, the subscript $k$ may be omitted. When $j$ is not specified, $j = 1$, corresponding to an observation while at the home node. For any target mask $t$, the depth range $D(t)$ and center $c^t$ can also be determined in the same manner as the palm and hand masks. The major axis of $t$ and change in depths for the pixels along this axis give the three components of the target orientation vector $\vec{o}$, or where necessary, $\vec{o'}_{k,j}$ or $\vec{o''}_{k,j}$.

Any edge can also be associated with a binary mask for the area swept through during motion along it, $s_{i,i'}$, approximated by a convex hull of the hand masks of the endpoint nodes, $h_i$ and $h_{i'}$. The depth range of motion along an edge is the full range between the minimum and maximum depths seen at either endpoint, $D(s_{i,i'}) \equiv [\textnormal{min}(D(h_i),D(h_{i'})), \textnormal{max}(D(h_i),D(h_{i'})]$. The direction of motion in image space along an edge $e_{i,i'}$ is described by $\vec{m}_{i,i'} \equiv c^p_{i'} - c^p_i$. Importantly, the direction of the final motion of a trajectory, which influences the likelihood of a successful grasp, is denoted $\vec{m}_{p,f}$. In particular, the agent will learn that $\vec{m}_{p,f}$ should align with $\vec{g}_p$ and $\vec{g}_f$, the orientations of the gripper fingers at nodes $n_p$ and $n_f$. As much as possible, these vectors should also be parallel to the direction from the penultimate and final node centers to the target center, $\vec{m}_{p,t} \equiv c^t - c^p_{p}$ and $\vec{m}_{f,t} \equiv c^t - c^p_{f}$, and perpendicular to $\vec{o}$ for the most reliable approach, as described in section \ref{sec:cos_sim_features}.

Finally, the agent may make a local, linear estimate of the relationship between the joint angles and the image of the hand in the neighborhood of a graph node. For $n_i$, this neighborhood is defined as $N(n_i) \equiv \{n_{i'} | \exists e_{i,i'}\}$. By the method discussed in section \ref{sec:local_jacobian}, comparing $q_i$ and $c^p_i$ with each $q_{i'}$ and $c^p_{i'}$ produces a local Jacobian estimate $\hat{J}(n_i)$ for the rate of change of image coordinates per change in joint angles. The inverse local Jacobian $\hat{J}^{-1}(n_i)$ may be used to estimate the joint angle perturbation necessary to move to desired image coordinates (such as $c^t$) rather than $c^p_{i}$. $\hat{J}^{-1}(n_f)$ is used for this purpose in section \ref{sec:local_jacobian}, and further used for preshaping in section \ref{sec:cos_sim_features}.

\section{Building a Representation of Peripersonal Space}

A baby begins to explore its environment and the range of motion of its arms with seemingly random movements and no clear external goal. Following this example, our agent begins with an empty PPS Graph $G$ and adds to it through observations of the motions it performs. At the beginning of the experiment, the end effector is placed near the center of the agent's field of view, with no joints near the extremes of their ranges. As each node $n_i$ is created, the images in $P(n_i)$ are analyzed to create the masks and vectors that represent the end effector state at $n_i$. In later learning of manipulation actions, the agent will be able to search these representations for specific features and values, mapping from a desirable stored image to the joint angles necessary to return to that state.

Each subsequent node is generated by making a small, random move from the previous node. This move is determined by a change in configuration of the arm. After recording $n_i$, $q_{i+1}$ is found by adding a small amount to each of the seven components of $q_i$. For each joint angle $k$, the displacement to add is sampled from a normal distribution with a standard deviation equal to a tenth of the full range of that joint. 
\begin{equation}
q^k_{i+1} = q^k_i + \Delta q^k \mbox{ where } \Delta q^k \sim N(0,\sigma_k) \mbox{ and } \sigma_k = 0.1 \cdot range(k)
\end{equation}

We impose a bias using a form of {\em rejection sampling}, requiring that the resulting end-effector pose must fall within the field of view, and must not collide either with the table or with the robot's own body.  If either condition is violated, the proposed configuration is rejected and a new $q_{i+1}$ is sampled.  As noted previously, human infants exhibit a bias toward keeping the hand visible \cite{vonHofsten-dp-82,vonHofsten-dp-84,vanderMeer-science-95,vanderMeer-ejpn-97}.  Human infants are also soft and robust, so they can detect and avoid collisions with minimal damage.  To prevent damage to our Baxter Research Robot, we implement these checks using a manufacturer-provided forward kinematics model that is below the level of detail of our model, and is used nowhere else in its implementation.  In future work, we will considering biasing this sampling to resemble human infants' pre-reaching motions toward objects, or to move in a cyclic fashion, often returning to the center of the field of view.

Once $q_{i+1}$ has been validated, the agent uses linear interpolation to move to this configuration from $q_i$. After arriving, the percepts for $n_{i+1}$ are recorded. An undirected edge $e_{i,i+1}$ is added to connect $n_i$ and $n_{i+1}$, as this motion can be proven to be safe to execute by construction. The visual representation of this edge is derived from the percepts of its endpoint nodes and stored by $G$. The edge $e_{i,i+1}$ has length $\vert \vert q_{i+1} - q_i \vert \vert$, the Euclidean distance between the nodes in joint space. The process of generating and recording new nodes is repeated until $n_{3000}$ and $e_{2999,3000}$ have been added to the graph.

\begin{figure}
\hrule
\vspace{1mm}
\begin{center}
\includegraphics[width=\textwidth]{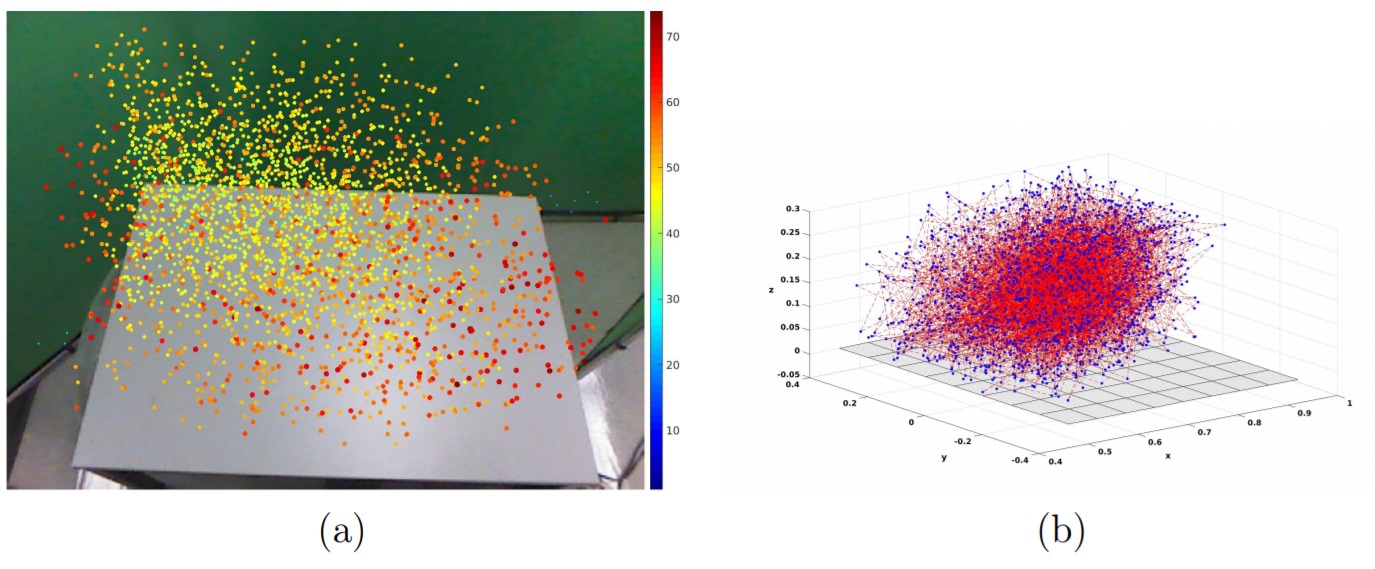}
\end{center}
\vspace{-4mm}
\caption{Two visualizations of the Peripersonal Space (PPS) Graph $G$. Each shows wide coverage that facilitates movement throughout the environment, but a few regions of relative sparsity, which may explain the increased difficulty of acting in those spaces. \textbf{(a)} An example RGB percept overlayed with the $(u,v)$ center of mass location for all 3000 nodes. The node markers are colored and sized according to mean disparity value $d$. Nodes with higher disparity (closer to the camera) appear larger and more red, while nodes with lower disparity (farther from the camera) appear smaller and more blue. \textbf{(b)} A representation of the graph that is unavailable to the agent. Nodes are plotted as blue points at their true world $(x,y,z)$ coordinates in the Baxter Robot's default frame of reference. The 2999 edges within the original chain from motor babbling are shown as dotted red lines, and the edges added according to the feasible length threshold are not displayed. The gray plane represents the surface of the table in the workspace, which is generally well-covered by $G$ except at the corner opposite the natural position of the agent's left hand.}
\label{fig:GraphCoverage}
\vspace{1mm}
\hrule
\end{figure}

At this point, $G$ is a sparse approximation of the configuration space of the robot arm, with enough nodes that the random motion has distributed them throughout the workspace. The overall coverage of the workspace is illustrated in Figure~\ref{fig:GraphCoverage}. However, the graph is a chain, and trajectories along edges could involve a very large number of moves.
In addition to inefficiency, having a single path through the graph will not provide options for avoiding obstacles or selecting the most reliable approach for a learned action. The graph should have much higher connectivity. To avoid making test motions between all pairs of edges, each possible edge is evaluated for feasibility by comparison to the initial set of edges from motor babbling. If the distance between two nodes is less than the mean length of all initial edges, an edge between them is added to the graph. Requiring a shorter distance to travel than the average edge known to be safe, the agent assumes these motions will also be feasible. With the inclusion of these edges, $G$ is complete and supports planning of multiple trajectories between pairs of nodes. A node where the arm rests naturally and that allows relatively unoccluded observation of the environment is designated as the \textit{home node}, $n_h$, where trajectories are planned and begin execution. Because $G$ is still a sparse approximation to the configuration space, trajectories across the environment will tend to be jerky.

\section{Learning a Reliable Reach Action}
\subsection{Observing the Unusual Event of a Bump}

During the construction of the PPS Graph, the agent's attention is focused on the appearance of the gripper and the arm. These regions of the image and their properties are highly variable relative to the background, making a clear and interesting violation of an initial static world model. In our case, the background fits the static world model since the early exploration was performed with no mobile objects present to simplify processing of the node images, and the magnitude of noise in the observations is small. More generally, with additional foreground objects present, portions of the image corresponding to the self will stand out due to the frequency of their movements and the correlation of these movements with motor signals.  Occasional change to other foreground objects can be left initially unmodeled, treated as noise in the visual percept.

After the PPS Graph is completely constructed, we then place several new objects in the agent's environment. All objects used for this work are rectangular prism blocks with a single long dimension, though the agent is not given this geometry. The blocks are placed upright at randomly generated coordinates on a table in front of the robot, with the requirement that each placement leaves all blocks fully within the field of vision and not occluded. We assume that the agent is capable of creating a binary mask for the location of an object in an RGB image. This capability currently relies on the objects having distinctive colors not present in the background so that masks can simply be fit to connected components of distinctively colored pixels, but this requirement could be relaxed with a more sophisticated image processing method. The agent uses the mask and accompanying depth image to determine the range of depths occupied by the object.

The agent continues to practice its new capability to perform motions allowed by the PPS Graph and observe the results of these motions. The agent follows this procedure:

\begin{enumerate}
\item{Observe the environment while at the home node, and find the initial mask for each of three movable objects placed in the foreground.}
\item{Select a random final node $n_f$ in the PPS Graph.}
\item{Perform a graph search to determine the shortest path trajectory from the home node to $n_f$.}
\item{Execute the trajectory, checking the visual percept at each node for any significant change to a pixel within an initial object mask.}
\item{If a change is observed or the current node is $n_f$, immediately return to the home node along the shortest path.}
\item{Find a final mask for each object, and calculate the Intersection over Union (IOU) between the initial and final masks, then cluster all IOU values seen so far into two clusters.}
\item{Repeat until the smaller cluster contains at least 20 examples.}
\end{enumerate}

We chose to place multiple foreground objects before each trajectory so that the agent could gather examples more efficiently than with a single object. The agent moved along 102 trajectories and gathered 306 IOU values while following this procedure. Of the 306 examples, most values are approximately 1, indicating a near match between the initial and final masks, but 21 are placed in a cluster of much lower IOU values. These numbers fit the intuition that a random trajectory would be unlikely to interact with a relatively small object. However, in a rare event the hand collides with the object, sliding it along the table or knocking it over, sometimes off the table (the resulting absence of a final mask leads to an IOU of 0, so no special case is necessary).

The agent can classify all future motions in the presence of an object by associating the observed IOU with one of the two clusters. While we human observers can describe the smaller cluster as a {\em bump} event, the robot learning agent knows only that the smaller cluster represents an unusual but recognizable event, worth further exploration.  By returning to the home node to observe the final mask, the agent can rule out an occlusion by the hand as the source of the change, and has not been observed to make false positive bump classifications. This is important so that the agent will not learn incorrect conditions for a bump. There are a small number of false negatives where the hand and object collide without lowering the IOU enough to fall into the rare cluster. The agent is still able to learn the conditions from the reduced number of observed bumps, and may even favor actions that cause larger, more reliable bumps as a result.

\subsection{Baseline Reaching with Random Trajectories}

We define a {\em reach} as a trajectory resulting in a bump event with a target object. The agent has learned how to evaluate a reach attempt according to the IOU clustering criterion, but has no knowledge of what makes a reach succeed. The current knowledge state cannot differentiate between good and poor trajectories, and sees each as having an equal, small chance of causing a bump. In the remainder of this section, the agent will consider additional features to plan increasingly reliable reach trajectories. In order to demonstrate this improvement, a baseline level of performance is computed over 40 trials with a single, randomly-placed target block and the current best policy of using the shortest PPS Graph path trajectory to a random final node. It is possible for an occlusion to cause a change within the initial object mask at a node partway through the trajectory, and the agent will return to the home node to observe the final mask as in the previous procedure. The agent should not allow this occlusion to prevent observation of a bump later in the trajectory. Therefore, if the IOU is an example of the usual case, the agent repeats the trajectory up to where it left off, ignoring any perceived changes to the initial mask, and then continues with the remaining moves to the final node, checking normally. This ensures that a bump anywhere along the full planned trajectory may be observed and the reach will be counted as successful. This method gives a baseline of 20\% reliability for the reach action, which we compare to the other methods in Figure~\ref{fig:ReachResult}.

\subsection{Identifying Reliable Final Node Candidates}\label{sec:candidates}

The agent has identified the rare event of a bump and the reach action which can cause this event, and is now intrinsically motivated to make this action more reliable. This will require identification of at least one feature that discriminates between the cases so far that have, or have not, resulted in a bump. The agent has a stored visual percept for each node in the PPS Graph and a current visual percept of the target object, and has extracted the masks and the depth ranges from each, so a comparison of these representations is straightforward.  All visual percepts have the same frame of reference, so any nonempty intersection corresponds to occupying the same region of the RGB image or the same depth, or both. Intuitively, a plan to move any part of the hand to occupy space already occupied by the block will instead push the block out of the way, causing the bump event. 

\begin{figure}
\hrule
\vspace{1mm}
\begin{center}
\includegraphics[width=\textwidth]{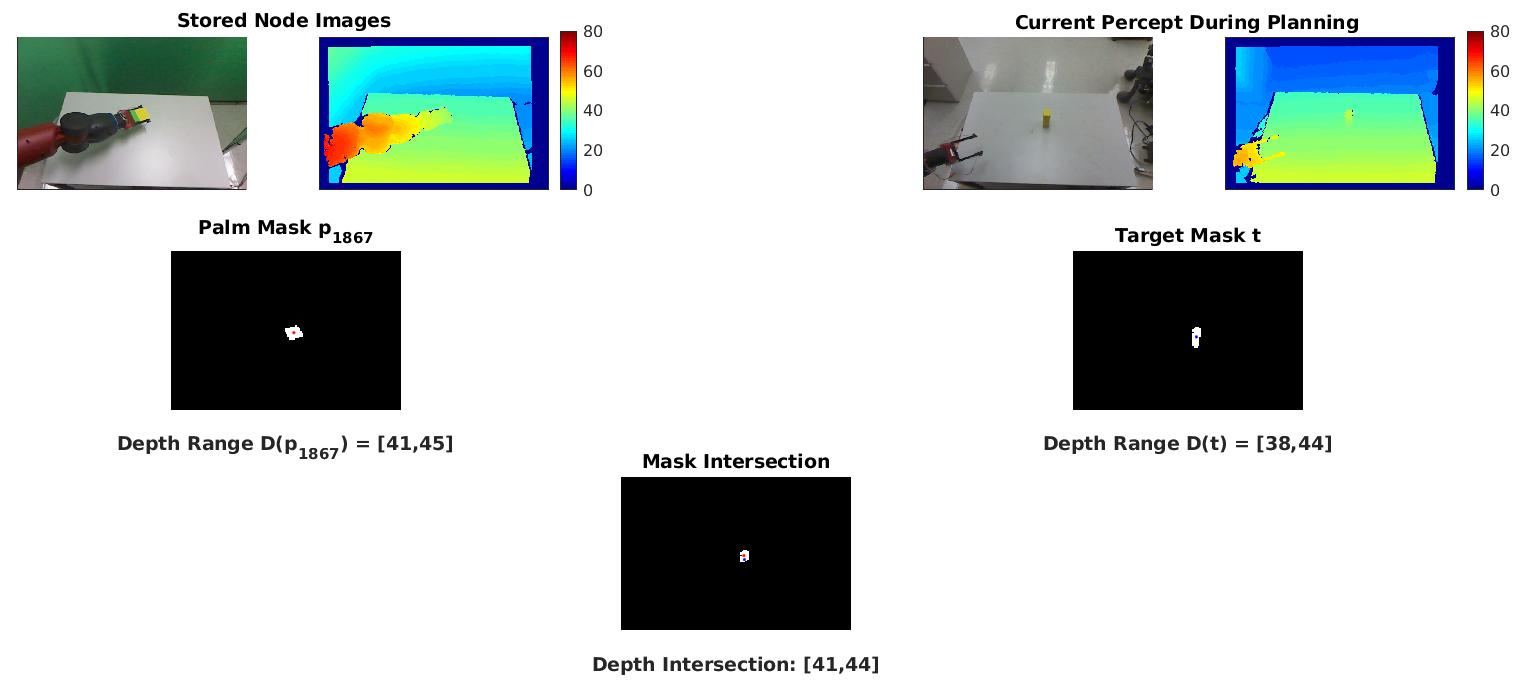}
\end{center}
\vspace{-4mm}
\caption{The process of identifying a candidate final node with favorable intersection features. The stored RGB-D percept taken upon first visiting the node (upper left three images) is processed to create its palm mask and depth range. The current percept of the environment (upper right three images) allows identification of a target object and creation of its mask and depth range. Eligibility to be a candidate depends on the presence of nonempty intersections of the masks and depth ranges (lower middle image). Both intersections are nonempty for this node, and it is recorded as a candidate. Also note the close proximity of the centers of the palm and target, which is learned as an additional feature for planning reliable reaches in Section~\ref{sec:nearest}. All 3000 nodes may be quickly evaluated as shown to generate the set of candidate final nodes.
}
\label{fig:candidate}
\vspace{1mm}
\hrule
\end{figure}

Without this intuition, the agent can discover this relationship in the data from the first 102 trajectories. For each binary mask $b \in \{p_f, h_f, s_{p,f}\}$ representing the hand at its final pose or throughout its final motion, and mask $t$ representing the target object, these examples are placed in four groups according to 
whether $b \cap t$ and/or $D(b) \cap D(t)$ are empty or nonempty. Counts of observed bumps and the total number of trajectories within each group allow the conditional probabilities of a bump to be computed. The set of groups where $b = p_f$ contains the group with the highest conditional probability. A bump is most likely (64\%) to occur at a node whose palm has a nonempty intersection in both mask and depth range with the target, that is, where 
\begin{equation}
p_f \cap t \neq \emptyset \;\wedge\; D(p_f) \cap D(t) \neq \emptyset.
\label{eqn:candidates}
\end{equation}
The set of nodes with these qualities are the best {\em candidate final nodes} for a reach trajectory. The process of identifying a node as a candidate is demonstrated in Figure~\ref{fig:candidate}. If no nodes meet both criteria, the set of candidates for $n_f$ will be extended until it is nonempty, first to include all $n_f$ such that $D(p_f) \cap D(t) \neq \emptyset$ (14\% bump likelihood), then all other nodes (1\% bump likelihood).

For environments with obstacles that should not be bumped, it may be necessary for the agent to plot courses that avoid additional bumps on the way to the target. In a setting with only one object, it is more efficient to avoid interactions with the target before the final node -- the agent repeats fewer motions after occlusions, and with fewer early bumps it misses fewer opportunities to gather information about the result of using the desired final node. In either case, a bump has never been observed when $s_{p,f} \cap t = \emptyset \wedge D(s_{p,f}) \cap D(t) = \emptyset$, so all but the final edge should have this property in order to have the lowest chance of an early bump.

This knowledge allows our agent to construct an improved reach action policy. The agent identifies the set of final node candidates with the best available intersection class, and then chooses a final node from this set randomly. The shortest graph path from the home node to this final node is found. If any nonfinal edges are perceived to have mask or depth range intersections with the target, they are temporarily removed, and the agent finds the shortest graph path without them. If no path remains, the process repeats, beginning with all edges allowed and ruling out only the edges that fail the next less strict criteria. For the same 40 placements as the baseline, 39 have at least one node with mask and depth range intersections with the target, and the policy of moving to one of these nodes bumps the target 21 times. Attempting a reach to the placement where no node has both RGB and Depth intersections was not successful. Overall, the reach action is now 52.5\% reliable. The comparison in Figure~\ref{fig:ReachResult} shows reaching is now more than twice as reliable as the baseline action with random final nodes.

\subsection{Selecting the Nearest Final Node Candidate}\label{sec:nearest}

The previous section included the intuition that attempting to occupy the same space with the hand and target should always cause a bump. This appears to conflict with the result of a 52.5\% reliable reach action. We propose two explanations for this difference. First, the representation of the palm with the full depth range over the full mask overestimates the space filled, and the predicted collision may occur in the empty space. This becomes more likely when noise in the visual percepts causes the masks or depth ranges to be more inclusive. Second, all candidates have an equal chance of being chosen even if the intersections are very small. If the intersection is small the collision may also be small, so the resulting object displacement may not be visibly identifiable, causing a false negative bump. This accounted for three of the perceived failures.

\begin{figure}
\hrule
\vspace{1mm}
\begin{center}
\includegraphics[width=\textwidth]{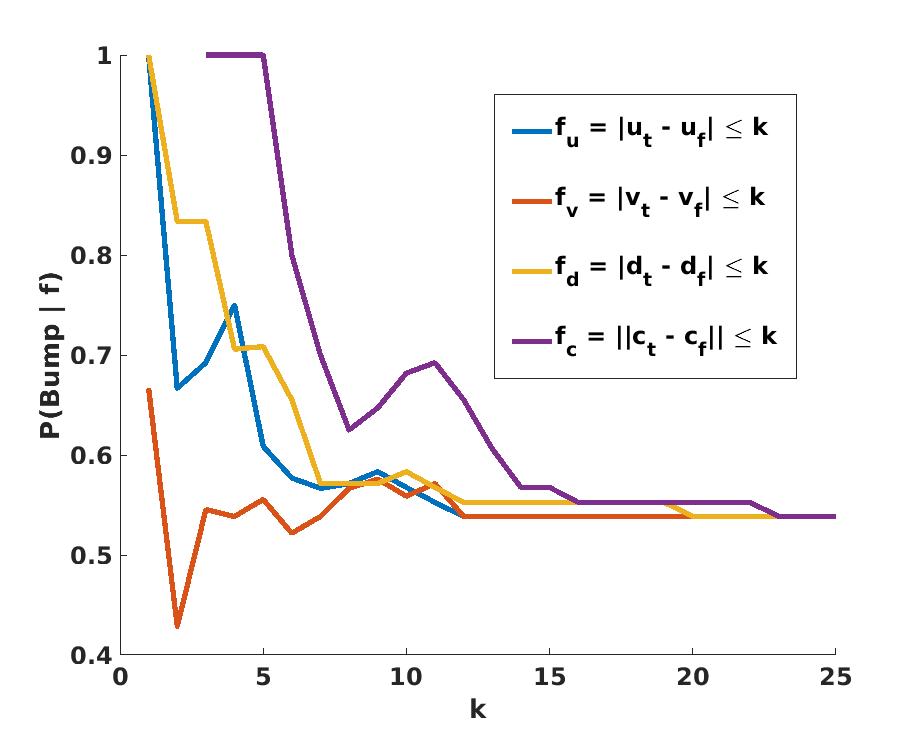}
\end{center}
\vspace{-4mm}
\caption{
The agent considers four possible features $f_u$, $f_v$, $f_d$, and $f_c$ for evaluating the conditional probability that moving to a node will cause a bump, given that the node is a candidate final node for which $f$ holds. These probabilities are based on the trajectories used in Section~\ref{sec:candidates} and their results. Each feature considers the centers of the palm and target in $(u,v,d)$ image-space. $f_u$, $f_v$, and $f_d$ evaluate to true if the absolute difference in one coordinate is less than a variable threshold $k$, and $f_c$ is true if the distance between centers is less than $k$. For all values of $k$, $P(Bump\mid f)$ is maximized when $f = f_c$. The agent chooses to evaluate all candidate final nodes with this feature. Reach trajectories are planned to end with the candidate that is nearest to the target, allowing the lowest threshold $k$, and therefore the highest probability of a bump.
}
\label{fig:By_Feature}
\vspace{1mm}
\hrule
\end{figure}

These problems illustrate that not all candidate final nodes are reliable for reaching, and in order to continue to improve, the agent must learn a method to replace its random selection among them. Features that describe the dependence of the probability of a bump on the difference between center positions of the palm and target are compared in Figure~\ref{fig:By_Feature}. This analysis supports the use of the final node candidate with the smallest center to center distance with the target $\vert \vert c^t - c^p_f \vert \vert$. Once this node is chosen, the rest of the path is planned as before. Attempting the 40 reaches again, the agent now considers the reach action to be 77.5\% reliable, with 31 successes, 7 false negatives, and 2 actual failures to bump the object. This result is also included in the comparison in Figure~\ref{fig:ReachResult}.

\subsection{Hill Climbing to the best Reaching Target Pose}\label{sec:local_jacobian}

Recall that the first improvement to the reach action was to identify a set of candidate final nodes, all nodes where the stored hand representation and the current percept of the target intersect in both the RGB and depth images. Moving to a random candidate final node instead of a fully random node more than doubles the rate at which bumps are successfully caused. However, Figure~\ref{fig:By_Feature} illustrates that not all candidate nodes are equally likely to produce a bump. Over the 40 trials of Section~\ref{sec:candidates}, the success rate for reaches decreased as $|| c^p_f - c^t ||$ increased. Choosing the nearest candidate improves the reliability of the reach to 77.5\%. However, this method for minimizing the distance between centers is limited by the density of the PPS Graph near the target. Especially in relatively sparse regions of the graph, even the nearest node may be insufficiently close for a reliable reach. The agent must learn to make small moves off the graph to reach closer to the object than the nearest node.

To make safe, useful moves to arm states that were not already explored and stored in $G$, the agent must extend its understanding of the arm's motion. The nodes of the PPS Graph provide a mapping between discrete arm configurations and the location and features of the hand in the corresponding stored images. The full model $J$ relating joint angle changes $\Delta q$ to changes in palm center coordinates $\Delta c$ is dependent on the current state of the arm $q$, a seven-dimensional vector. As it is also nonlinear, this model is prohibitively difficult to learn with the agent's current level of experience and features. However, $G$ contains sufficient data for making linear approximations of the relationship between $\Delta q$ and $\Delta c$ local to a particular $n_i$. This estimate is most accurate near the configuration $q_i$, with increasing error as the distance from $q_i$ increases.

\begin{figure}
\hrule
\vspace{1mm}
\begin{center}
\includegraphics[width=\textwidth]{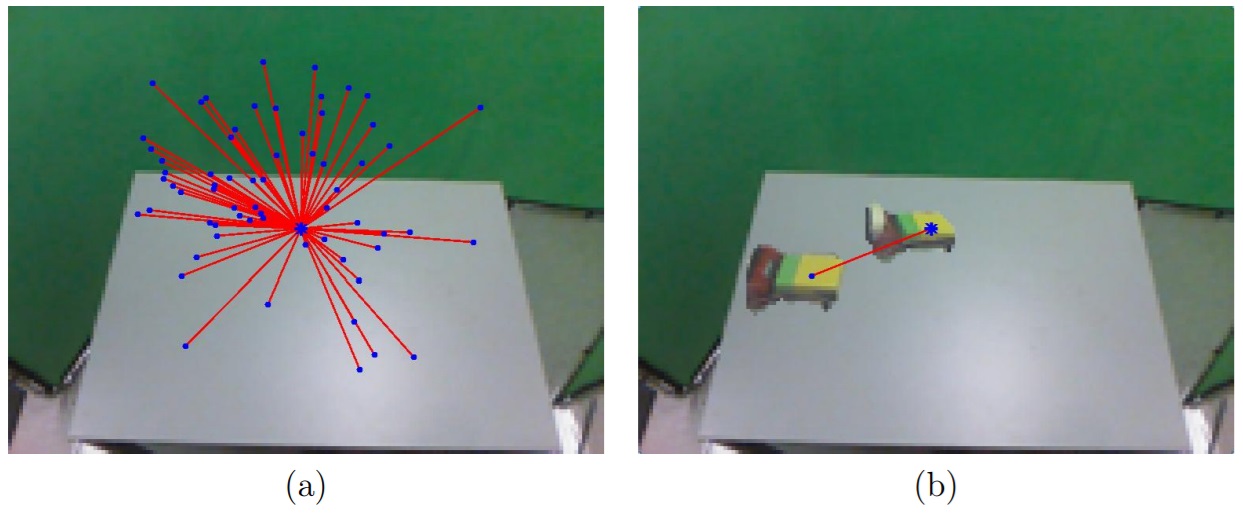}
\end{center}
\vspace{-4mm}
\caption{\textbf{(a)} The agent considers the graph neighborhood around a node $n_i$ to estimate the change in appearance for each small change in configuration near that node. The predictions will be made by a local Jacobian estimate $\hat{J}(n_i)$. $n_i$ (shown larger) is near the center of the graph and has a large number of neighbors. Each edge has a short length in configuration space, where edge feasibility is measured, even though some neighbors appear relatively distant in image space. The furthest neighbors tend to be those where almost all of the edge length comes from a difference in proximal joint angles that have a larger affect. The large number of edges provides more information for the local Jacobian estimate. \textbf{(b)} The images of the node and one of its neighbors are superimposed with a representation of the edge, drawn between their centers of mass. This provides one example of a change in configuration $\Delta q$ and the resulting change in center locations $\Delta c$. The examples for all edges may be combined and the best estimate for $\hat{J}(n_i)$ solved for. The inverse $\hat{J}^{-1}(n_i)$ may also be computed as a method for estimating the change $\Delta q$ necessary for a desired change $\Delta c$. Both estimates are accurate in practice near $n_i$, and adjustments to $q_f$ according to $\hat{J}^{-1}(n_f)$ make the reach action completely reliable.}
\label{fig:jacobian}
\vspace{1mm}
\hrule
\end{figure}

The local Jacobian estimate $\hat{J}(n_i)$ considers all edges $e_{i,i'}$ such that $n_{i'} \in N(n_i)$. Each edge provides an example pair of changes $\Delta q = q_{i'} - q_i$ and $\Delta c = c^p_{i'} - c^p_i$. If there are $m$ neighbors, and thus $m$ edges, these can be combined as an $m \times 7$ matrix $\Delta Q$ and a $m \times 3$ matrix $\Delta C$, respectively. $\hat{J}(n_i)$ is the least squares solution of 
\begin{equation} 
  \Delta Q \; \hat{J}(n_i) = \Delta C. 
\end{equation}
Figure~\ref{fig:jacobian} shows an example graph neighborhood and a visualization of the information contained in each edge used in this process. The resulting $\hat{J}(n_i)$ is a $7 \times 3$ matrix where the element at $ \left[row,col\right] $ gives the rate of change for $c^{col}$ (either the $u$, $v$, or $d$ coordinate of the palm's center of mass) for each unit change to $q^{row}$. A possible adjustment $\Delta q$ to $q_i$ may be evaluated by determining if the predicted new palm center $\hat{c}^p_i \equiv c^p_i + \Delta q\hat{J}(n_i)$ and $p_i$ translated by $\Delta q\hat{J}(n_i)$ have desirable features. Rotations and shape changes of $p_i$ that will occur during this motion are not modeled, but are typically small.

The agent may improve the accuracy of the final motion of its reach by changing the arm configuration of the final pose from $q_f$. The agent may generate and test $\Delta q$ values, looking for movements such that $\vert \vert c^t - \hat{c}^p_f  \vert \vert < \vert \vert c^t - c^p_f \vert \vert$. However, it is much more efficient to use the pseudo-inverse of the local Jacobian estimate, $\hat{J}^{-1}(n_f)$, which gives the approximate change in each joint angle for unit changes in the image coordinates of the center of mass. The agent can then solve for the ideal modified configuration $q^*_f$. $\vert \vert \hat{c}^p_f - c^t \vert \vert = 0$ is expected to be maximally reliable, so the change $\Delta c = c^t - c^p_f$ is desired. Multiplying $\Delta c$ by $\hat{J}^{-1}(n_f)$ gives a value for $\Delta q$ such that the joint angle changes are predicted to cause $\Delta c$. The ideal (estimated) final configuration $q^*_f$ is the sum of this $\Delta q$ and the original configuration $q_f$, or
\begin{equation}\label{eqn:final}
q^*_f = q_f + (c^t - c^p_f)\hat{J}^{-1}(n_f)
\end{equation}

When the agent moves to $q^*_f$, the palm should be placed so that its center matches the target's center, making a collision highly likely. Relying on intersections that occur near the boundaries of the binary masks or depth ranges to cause a bump is susceptible to noise in the image, as there may be no true intersection. It is also possible that the intersection will involve too little of an object to cause an observable bump from the collision. A motion that aligns the centers is robust to noise and should increase the size of the intersection and resulting bump.

\begin{figure}
\hrule
\vspace{1mm}
\begin{center}
\includegraphics[width=\textwidth]{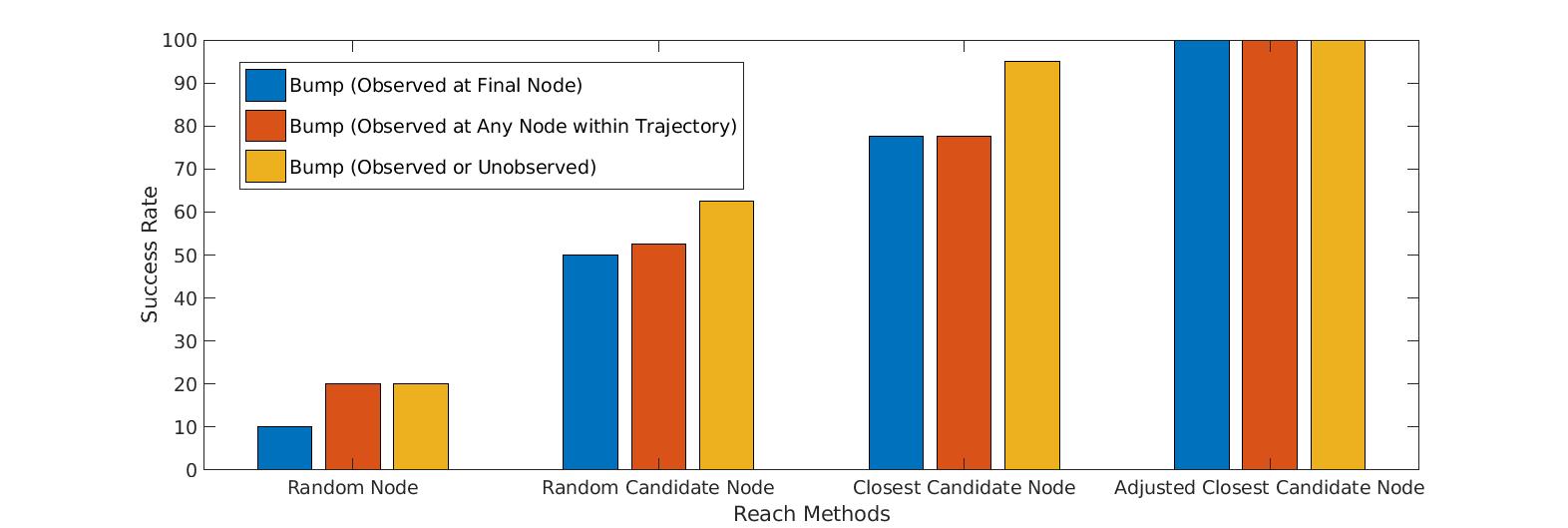}
\end{center}
\vspace{-4mm}
\caption{Reliability of the agent's action to reach and bump a single target object by following a trajectory to a selected target node. The four groups represent (1) randomly selected target node; (2) random selection from among candidate nodes with non-empty image intersections; (3) select closest among candidate nodes; (4) adjust node with local Jacobian to best match target object.  Within each group, the bars represent different criteria for success:  (l) observed bump at final node, which measures the agent's ability to cause bumps intentionally and efficiently; (m) observed bump anywhere in the trajectory, which identifies bumps that can be learned from; (r) any bump, observed or unobserved, which measures ground truth of bump occurrence.  
}
\label{fig:ReachResult}
\vspace{1mm}
\hrule
\end{figure}

While the ability to make a small move off of the graph to $q^*_f$ increases the robustness of the reach, it does not eliminate the need for a set of candidate final nodes and the decision to use the nearest one as $n_f$. As $\hat{J}^{-1}(n_f)$ is a local estimate, if $\vert \vert c^t - c^p_f \vert \vert$ is large, the error in the recommended $\Delta q$ will also tend to be large. Choosing the nearest candidate $n_f$ minimizes the factor by which natural errors in $\hat{J}^{-1}(n_f)$ will be multiplied, giving the best accuracy for the final position of the reach. Adding the use of the inverse local Jacobian gives the final reaching procedure below. Using this procedure on the training set of target placements, the agent perceives bumps at the final node of all 40 trajectories. This 100\% result demonstrates that the reach action has become reliable, and is a significant improvement from the previous methods shown in Figure~\ref{fig:ReachResult}.

\section{Learning a Reliable Grasp Action}

As the reach action toward a target object becomes more reliable, the result of causing a quasi-static change in the image of that object becomes more typical.  However, there is an unusual result:  the ``accidental grasp'', where the object happens to trigger the Palmar reflex, causing the grippers to close, and grasping the object.  This is a second instance of the intrinsic motivation pattern that drove learning to reach, and it can now drive the process of learning to grasp more reliably.

Eventually, this grasp action will serve as part of a pick and place operation in high level planning.

\subsection{Monitoring the Palmar Reflex During Reaching}

Before the agent can focus on learning to grasp, it is necessary to have a sufficient number of experiences of grasps as accidental and rare events. Without enough examples, learning the conditions for a grasp may prove too difficult, leading to a modest rate of improvement and a low reward. In our model, the agent focuses next on an intermediate rare event. A {\em Palmar bump} is a bump that also triggers the agent's Palmar reflex, causing the grippers to close. The set of trajectories that cause Palmar bumps is a superset of the set of trajectories that will cause grasps, since the agent must reach to the object and close its grippers on it to perform a grasp. Palmar bumps that do not result in a grasp can be seen as near successes, and the trajectories used share similar properties with those of successful grasps. As a result, gathering more examples of Palmar bumps, some of which should be grasps, will prepare the agent for its next goal of making the rare grasp event reliable.

The openness of the gripper is a degree of freedom for the robot's motion, and is continually sensed by proprioception. As a result, accurate detection of when the Palmar reflex has been triggered does not rely on the visual percept, and can be observed in a rapid decrease of openness to a new fixed point. In the agent's final set of 40 reach attempts, 12 Palmar bumps occurred. We show this result alongside the rest of the results for this section numerically in Figure~\ref{fig:GraspResult} and spatially in Figure~\ref{fig:training_set_positions}.

\subsection{Learning to Initiate Grasps with the Gripper Fully Open}

\begin{figure}
\hrule
\vspace{1mm}
\begin{center}
\includegraphics[width=\textwidth]{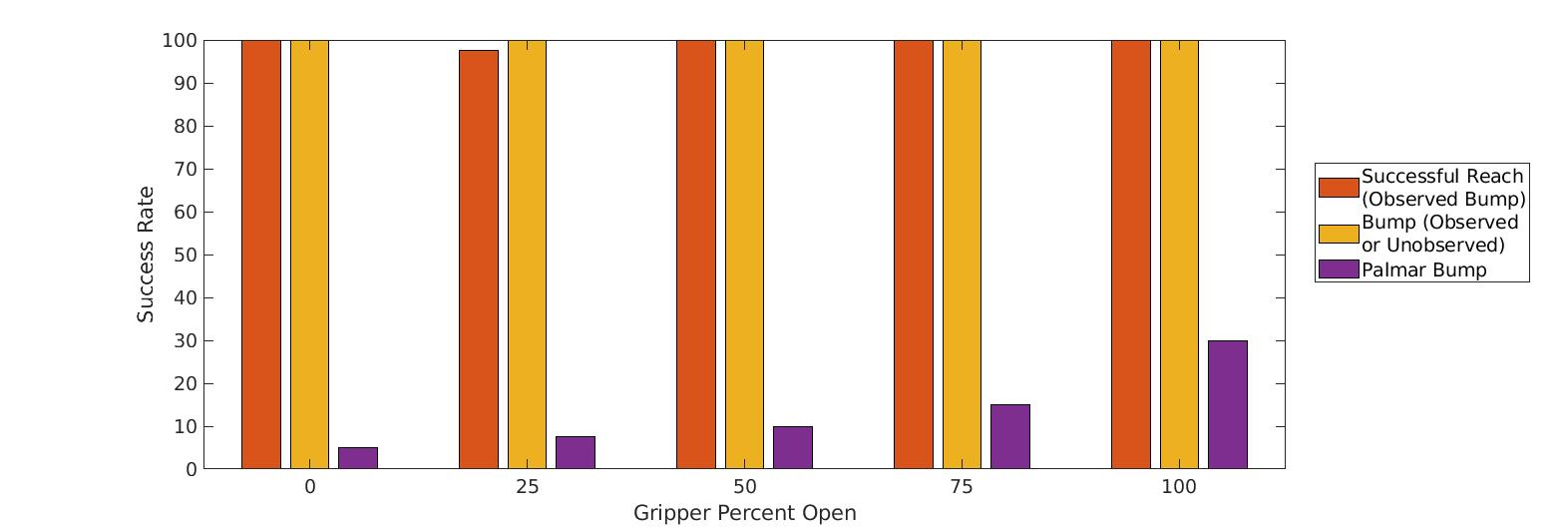}
\end{center}
\vspace{-4mm}
\caption{The portion of attempted reach trajectories that produce observed bumps (orange), ground truth bumps (yellow), and Palmar bumps, or bumps which also trigger the Palmar reflex (purple) for varying gripper apertures $a$. The high reliability of the reach action is independent of $a$, indicating it could be learned and executed with any setting. By contrast, triggering the Palmar reflex is much more likely as $a$ increases, and is learned as a prerequisite for the Palmar bump event and later for the grasp action.}
\label{fig:GripperOpen}
\vspace{1mm}
\hrule
\end{figure}

When the PPS Graph was created, the robot grasped a brightly colored block the full width of the grippers to aid in visual tracking. As a result, the default setting for the gripper is 100\% open at all nodes, and the agent has planned to use this setting for all motions so far, only experiencing different settings when the Palmar Reflex closes the grippers. While it is intuitively desirable for the agent to approach targets with the grippers open for a Palmar Bump or grasp, the agent does not yet have sufficient data to reach this conclusion. The agent repeats the final 40 reaching trajectories with the gripper 0\%, 25\%, 50\%, and 75\% open and compares these results with those it already has for 100\% open, as shown in Figure~\ref{fig:GripperOpen}.

Two conclusions may be drawn from the results of this experiment. First, it is clear that the probability of a Palmar Bump increases with the openness $a$ of the gripper during the approach. The agent will continue using the fully open setting in future attempts to maximize its expected success rate. Second, we see that the openness of the gripper has almost no affect the probability of a bump. In fact, only one trial was perceived to fail with any setting, and this was a false negative. We claim that this demonstrates the agent could have learned the reach action with the same process and ending reliability for any gripper setting, and at that point would learn to prefer 100\% open. It is therefore not necessary for our model to assume any initial setting $a$ for the gripper opening while learning to reach.

\subsection{Planning the Approach with Cosine Similarity Features}\label{sec:cos_sim_features}

When reaching, it is important that the candidate final nodes satisfying equation~(\ref{eqn:candidates}) are identified, and $n_f$ is chosen to minimize $\vert \vert c^t - c^p_f \vert \vert$. To plan reaches that cause Palmar bumps, additional features are needed to ensure not only that the final position is correct, but also that the hand orientation and the direction of final motion are suitable. The agent approximates the orientation of the hand throughout the final motion with the gripper vectors at its endpoints, $\vec{g}_p$ and $\vec{g}_f$. To carry out a reliable grasp, the agent must consider the relationship between these, the direction of the unmodified final motion $\vec{m}_{p,f}$, the direction the final motion must take to reach the target $\vec{m}_{p,t}$, and the direction of the displacement between the stored final pose and the target $\vec{m}_{f,t}$. Finally, the agent accounts for variations in the target placement by considering the object's orientation $\vec{o}$. 

To discover the best relationship between these vectors for repeating the Palmar bump event, the agent considers the cosine similarity $C(\vec{v}_1, \vec{v}_2)$ of each pair $\vec{v}_1, \vec{v}_2 \in \{\vec{g}_p,\vec{g}_f,\vec{m}_{p,f},\vec{m}_{p,t},\vec{m}_{f,t},\vec{o}\}$ in the data from the final reaching experiment (Section~\ref{sec:local_jacobian}). The cosine similarities are discretized to the nearest value in \{-1, -0.5, 0, 0.5, 1\}. The rate of Palmar bumps is observed for trajectories grouped by their discretized $C$ values. When $\vec{v}_1 \neq \vec{o}$ and $\vec{v}_2 \neq \vec{o}$, the highest rate of Palmar bumps occurs in the $C(\vec{v}_1, \vec{v}_2) \approx 1$ group. For any $\vec{v}_1 \neq \vec{o}$, the trajectories where $C(\vec{v}_1, \vec{o}) \approx 0$ have the highest rate. The agent concludes that the ideal approach for the Palmar bump event should use matching directions for all vectors describing the motion and orientation of the hand, $\{\vec{g}_p,\vec{g}_f,\vec{m}_{p,f},\vec{m}_{p,t},\vec{m}_{f,t}\}$, and all of these parallel vectors should be perpendicular to the target's major axis $\vec{o}$.  

\begin{figure}
\hrule
\vspace{1mm}
\begin{center}
\includegraphics[width=\textwidth]{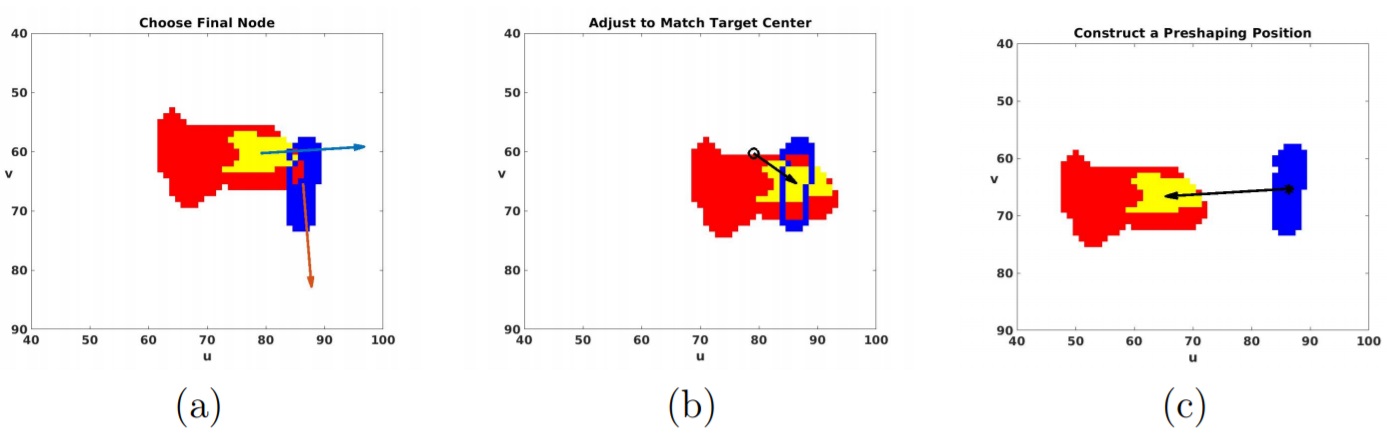}
\end{center}
\vspace{-4mm}
\caption{The agent plans modifications to the end of the trajectory, and defines a preshaping position.  In particular, it identifies cosine similarity features among vectors characterizing the hand-object configuration that maximize the probability of a successful grasp.  Human intuition recognizes that the effect of this optimization is that all vectors describing gripper direction and the direction of motion should be near parallel, with all of these vectors near perpendicular to the target's major axis. \textbf{(a)} The agent first identifies the set of candidate final nodes, nodes for which the stored mask and depth range intersect with those of the target. From this set, it chooses $n_f$ to minimize $C(\vec{g}_f,\vec{o})$. This image displays the palm mask (yellow) and hand mask (red) for the chosen $n_f$, along with the target mask (blue). A blue outline is used to show the boundary of the intersection between the hand and target. $\vec{g}_f$ and $\vec{o}$ are displayed in light blue and orange, respectively. \textbf{(b)} The agent uses $\hat{J}^{-1}(n_f)$ to estimate the change in joint angles necessary to cause the image-space translation shown here. When this translation is used, the agent aligns $c^p_f$ with $c^t$ by moving to $q^*_f$.  The agent chooses to make this adjustment as it was found that minimizing $\vert \vert c^p_f - c^t \vert \vert$ maximizes the reliability of the reach portion of the grasp. We can see that this improves the accuracy of the grasp approach motion toward the object. \textbf{(c)} Our understanding of the grasp action suggests that leading the motion with the opening between the grippers allows the grippers to surround and grasp the object before it can be bumped away by the outside of the hand. The agent has learned that it is best to align $\vec{g}_f$, $\vec{g}_p$, and $\vec{m}_{pt}$ (i.e., to maximize their pairwise cosine similarities), which produces this behavior. To maximize these similarities, the agent can use the local Jacobian to find a configuration $q^*_p$ that displaces the hand from $q^*_f$ in the opposite direction of $\vec{g}_f$. This modification should keep $\vec{g}^*_p \approx \vec{g}_f$. Therefore, using $q^*_p$ instead of $q_p$ during the final motion gives the three vectors close to the same direction.
}
\label{fig:cos_sim_figure}
\vspace{1mm}
\hrule
\end{figure}

The agent will use these conclusions to plan the next set of trajectories to interact with the target. At this time, the agent does not have the ability to change any $\vec{g}_i$ to a particular direction to be perpendicular to $\vec{o}$. Therefore, instead of the nearest candidate final node, $n_f$ is selected from the candidates such that $\vert C(\vec{g}_f,\vec{o}) \vert$ is minimized. As before, $\hat{J}^{-1}(n_f)$ is computed and used to modify the final configuration to a more reliable $q^*_f$. The agent may apply $\hat{J}^{-1}(n_f)$ again to create a preshaping position, a copy of the final position translated in the direction of $-\vec{g}_f$. This image-space translation has a magnitude of 21, the mean length of the final motion for all Palmar bumps previously observed. The preshaping position has configuration $q^*_p$, and will replace $q_p$. With this use of $\hat{J}^{-1}(n_f)$, it is expected that $\vec{g}_p \approx \vec{g}_f$, and the motion from $q^*_p$ to $q^*_f$ should be in the direction of $\vec{g}_f$, opposite of the translation. In place of $\vec{m}_{p,f}$, $\vec{m}_{p,t}$, and $\vec{m}_{f,t}$, the direction of this motion is parallel to the gripper vector and near perpendicular to the target major axis. The three steps of choosing $n_f$, adjusting to $q^*_f$ to match centers with the target, and translating to create a well-aligned preshaping position are visualized in Figure~\ref{fig:cos_sim_figure}.

The agent must plan a trajectory that ends with this approach. $q^*_p$ is not stored in $G$, so to find a feasible path to $q^*_p$, the agent first identifies the nearest node $n_n \in G$ that minimizes $\vert \vert q^*_p - q_n \vert \vert$. A graph search then yields the shortest path from the home node to $n_n$. After visiting $n_n$, the arm will be moved from $q_n$ to $q^*_p$, and then make the final motion to $q^*_f$ to complete the trajectory. Using trajectories planned in this manner on the training set, 39 of 40 reaches are successfully completed, and 21 of these cause Palmar bumps, 14 of which grasp the target. Figures~\ref{fig:GraspResult} and \ref{fig:training_set_positions} provide comparisons with results from other learning stages.

\subsection{Observing the Unusual Event of a Grasp}

In order to learn a new action, intrinsic motivation drives the agent to investigate and repeat an unusual event it has observed. The reach action is learned to enable reliably changing the target object to a new state, observed visually. A subset of these bumps are accompanied by a proprioceptive sensation that the grippers close as the Palmar reflex is triggered, changing the value of $a$. With enough experience from attempting to produce Palmar bumps, a subset of these events, grasps, can be identified by a qualitative difference in the visual state of the target. Some Palmar bumps appear similar to the results of a typical bump, with the target moving quasi-statically to a new static pose. For Palmar bumps that are also grasps, the target object exhibits a sequence of new states, following the motion of the hand.

\begin{figure}
\hrule
\vspace{1mm}
\begin{center}
\includegraphics[width=\textwidth]{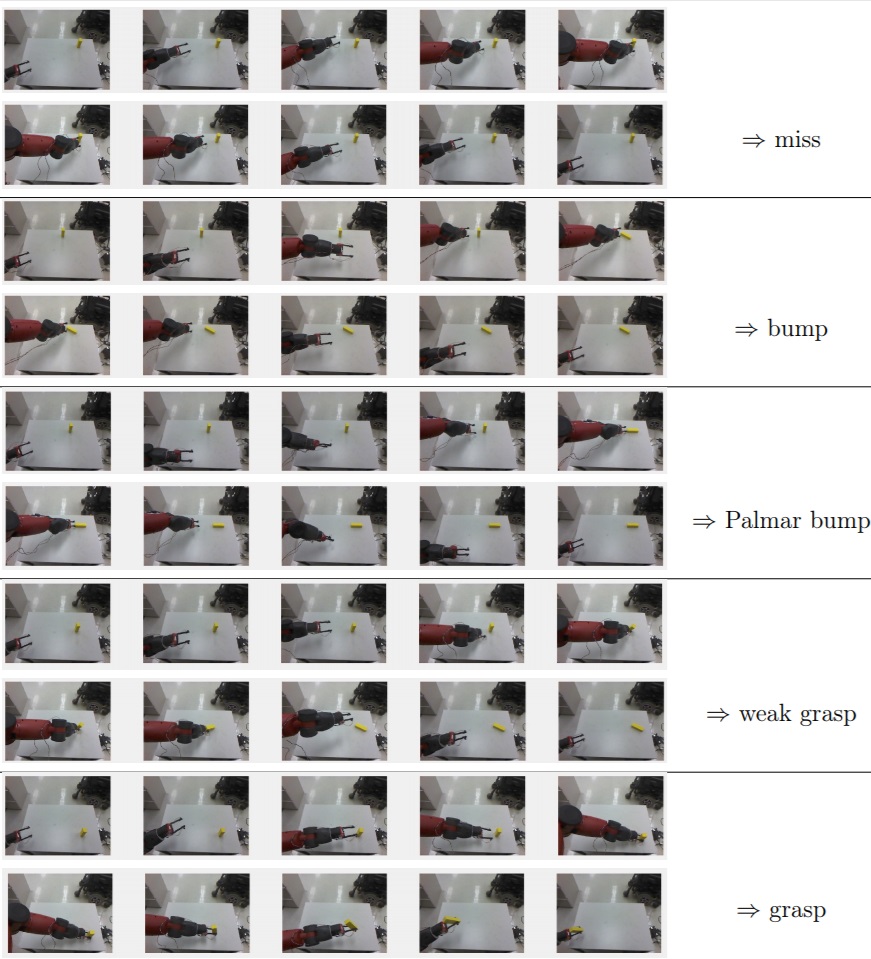}
\end{center}
\vspace{-9mm}
\caption{The agent's RGB percepts during attempted grasp trajectories. Images for the forward portion toward the final node, $\forall j P_{RGB}'(n_{T_j})$, are shown in the first of each pair of rows, and images for the portion to return to the home node, $\forall j P_{RGB}''(n_{T_j})$, are shown in the second rows. For trajectories with $\vert T \vert > 5$, images for some nodes in the middle of the trajectory have been omitted. The agent classifies the result of the grasp attempt by observing the state of the target object during the trajectory. If there is no change between the first and last observations, the object has been missed entirely. In all other cases these observations should be significantly different, and the reach component of the grasp was successful. Further classification depends on the state throughout the return trajectory. When the object takes a new quasi-static position and shows no further change, this is a bump that does not grasp, or a Palmar bump that does not grasp if the Palmar reflex was also triggered. Among the bumps and Palmar bumps, a subset with more than one observed change are also grasps. If the object has continued motion corresponding to the motion of the hand, a grasp has been initiated. If this quality of following the hand persists through all of $P_{RGB}''(n_{T_j})$, the grasp is successful. Otherwise the grasp is considered weak, and is not counted among the successes.}
\label{fig:trajectory_results}
\vspace{1mm}
\hrule
\end{figure}

The agent can identify this corresponding motion by comparing masks and depth ranges during the return trajectory. A grasp is successful if and only if 
\begin{equation}
\forall j \; [ \vert h_{T_j} \cap t''_j \vert > 0 \;\wedge\; \vert D(h_{T_j}) \cap D(t''_j) \vert > 0 ].
\end{equation}
That is, if the stored masks and depth ranges for each node of the trajectory intersect with those of the target object in the current visual percept, the object is considered to be moving with the hand after a successful grasp. Note that the full hand masks and depth ranges are used since the gripper fingers, once closed, may obscure the portion of the object in the palm region. If the comparison for at least one mode does not show an intersection, the grasp is considered unsuccessful, either because the initial interaction failed to grasp the object or the grasp was poor and did not persist through the return trajectory. By considering both situations to be failures, grasps that are successfully learned are more likely, in subsequent learning, to facilitate higher order actions that would use this grasp. Example trajectory images and classifications are provided in Figure~\ref{fig:trajectory_results}.

\subsection{Learning to Orient the Wrist from Past Successes}\label{sec:wrist}

By this time, the agent has observed that, like the gripper aperture $a$, $q^7$ (the angle for Baxter's $w2$ joint, the most distal twist joint) does not have a significant impact on the hand's location in the image, and may be varied freely while still being considered to visit a node.  Rotating $w2$ affects only a small portion of the wrist and applies a roll relative to the axis of the wrist and forearm. This alters the orientation and perceived shape of the gripper opening, but leaves the position largely unchanged.

This freedom allows the agent to explore grasps with different $q^7$ settings in an attempt to place the open grippers around the object along a narrow dimension. Doing so does not conflict with the previous method of choosing $n_f$ such that $\vec{g}_f$ and $\vec{o}$ are approximately parallel, as the roll of the hand is varied without changing the axis of the forearm. In order to avoid failures from large, sudden rotations of the hand near the target, when a new $q^7$ is chosen it will be used instead of the stored $q^7$ value of all nodes in the trajectory $n_{T_j}$.

To begin, the agent repeats each successful grasp, with a linear search over values of $q^7$ to identify the longest continuous range where the attempt still succeeds. The center of this range will be saved as the ideal $q^7$ value for this example grasp. To attempt a new grasp, the adjusted final configuration $q^*_f$ is computed by equation~(\ref{eqn:final}). Using the Euclidean distance between all other joint angles, $\{q^1_f, ..., q^6_f\}$, the nearest neighbor example grasp is found for the current trial. The grasp is attempted with the ideal $q^7$ value from this example and all other angles unchanged.

Over the same training set of 40 object placements, this technique increases the number of Palmar bumps to 30 (75\%), and grasps to 20 (50\%), as shown in Figures~\ref{fig:GraspResult} and \ref{fig:training_set_positions}. These increases come at the cost of one bump, where the target is now missed because the rotation of the hand prevents a collision that used to narrowly occur. In principle, any time new successes are achieved, they can be treated as new example grasps with ideal $q^7$ values to consider for trials with nearby target placements, allowing for further improvements to the success rate. However, in this training set only two still unsuccessful grasp attempts have different nearest neighbor examples than previously, and neither changes to a success with the new $q^7$ value. Iterations of using new nearest neighbors therefore end, but may be returned to in future work once more examples are available.

\subsection{Fine-tuning to Improve Grasp Reliability}\label{subsection:fine_tuning}

At this point, the agent observes successful grasps several times more often than occurred accidentally during reaching. This experience also provides information on why the other attempts fail, especially near misses where the error can be well explained by a single factor. In future work, these factors could be better observed by processing images or video during the grasp attempt, or with tactile sensing, and the agent could learn to make its own corrections. In the remainder of this work, we implement the kinds of intuitive adjustments to the grasp method that we anticipate being learned autonomously in future work.  We present the resulting reliability of the grasp action in this section.

With the shift of focus from precise positioning to approach angle and wrist orientation, the agent has begun to sometimes fail the reach portion of the grasp, missing the object entirely and making a grasp impossible. Recalling that the closer $c^p_f$ is to $c^t$ before adjustment, the more likely a bump is to occur, it can be further observed that all successful grasps have had a final motion with length $\vert \vert c^t - c^p_f \vert \vert < 21$. To reduce the number of misses and to improve the accuracy of the inverse local Jacobian estimates, all final node candidates where $\vert \vert c^t - c^p_f \vert \vert \geq 21$ are no longer considered.

Additional failures occurred because the agent completed the final motion rather than stopping the arm after the Palmar reflex was triggered. In some trials, the continued motion knocked the object out from between the gripper fingers, causing a Palmar bump without a grasp. In two instances, the grasp was completed, but the attempt to carry through the full motion pressed the object into the table surface, causing it to slip. This weakened the grasp enough that it failed before the return in completed. Our agent cannot currently sense the additional force from the table or the slipping of the object, but can implement the correction to stop immediately when the Palmar reflex is triggered. We believe it is natural that the reflex event would both start the closing of the grippers and end the motion of the arm. This change to the grasping procedure converted one Palmar bump and both weak grasp trials to successful grasps.

\begin{figure}
\hrule
\vspace{1mm}
\begin{center}
\includegraphics[width=\textwidth]{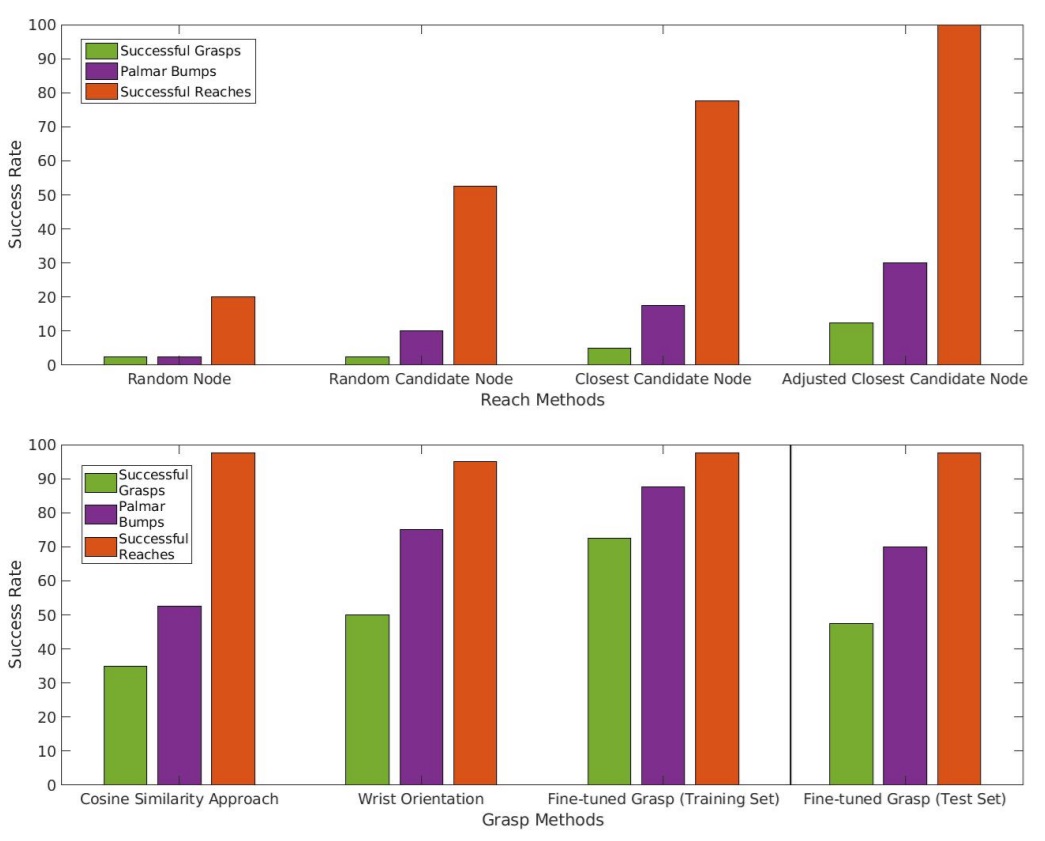}
\end{center}
\vspace{-4mm}
\caption{The top plot presents the overall results from the reaching methods as a baseline for the grasp action. The final reach method, Adjusted Closest Candidate Node (section \ref{sec:local_jacobian}), is always successful at reaching, but within these interactions only 12.5\% are fully successful though accidental grasps. By considering additional features, the grasp methods in the bottom plot all achieve more than double this success rate for grasping with only modest decreases in reach reliability. The Cosine Similarity Approach Method (section \ref{sec:cos_sim_features}) aims to increase the number of Palmar Bumps, with $n_f$ chosen from the candidates such that $\vert C(\vec{g}_f,\vec{o}) \vert$ is minimized and with $n_p$ replaced by a preshaping position so that all other cosine similarities are 1. Approaching with a motion parallel to $\vec{g}_f$ and perpendicular to $\vec{o}$ also increases the number of successful grasps. The Wrist Orientation Method (section \ref{sec:wrist}) further adds a technique to copy the most distal degree of freedom $q_7$ used at the nearest configuration to previously succeed, converting more bumps into Palmar bumps and grasps. Finally, correcting for perception and motion errors with the Fine-tuned Grasp Method (section \ref{subsection:fine_tuning}) yields the best overall result of 72.5\% grasping and 97.5\% reaching. When evaluated on previously unseen positions, the grasp success rate decreases to 47.5\%. The similarity to the training set result before fine-tuning suggests that the corrective techniques may be overfit, but much of the grasp skill gained has generalized. For example, the subset of learned skills needed for the reach action applies well to new target placements, with an equal 97.5\% of grasp attempts at least bumping the target in both the training and test sets.}
\label{fig:GraspResult}
\vspace{1mm}
\hrule
\end{figure}

\begin{figure}
\hrule
\vspace{1mm}
\begin{center}
\includegraphics[width=\textwidth]{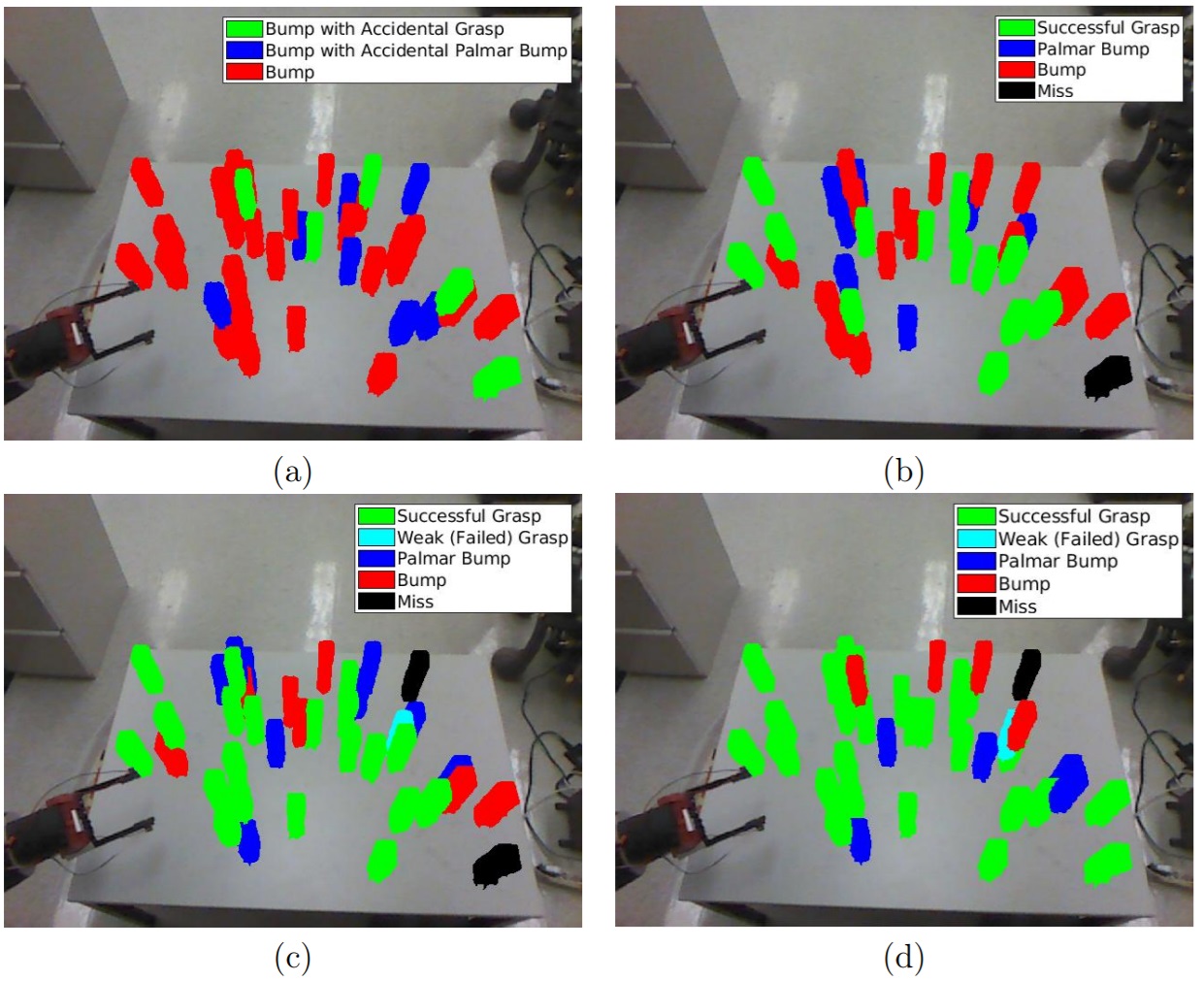}
\end{center}
\vspace{-4mm}
\caption{Spatial representations of the results of four methods for the agent's learned reach and grasp actions. Each shows a superposition of all placements of the single target object, colored according to the result of the agent's attempt to repeat an unusual event by executing a motion trajectory. When two classifications apply, the more specific one is used (e.g. a grasp is also a bump, but is shown as a grasp only). \textbf{(a)} The final reaching method (Section~\ref{sec:local_jacobian}) successfully repeats the bump event for all target placements. A small number of these reaches accidentally trigger the Palmar reflex, five of which become early examples of grasps. \textbf{(b)} Using cosine similarity features (Section~\ref{sec:cos_sim_features}), the agent modifies the final approach so that this motion causes significantly more Palmar bumps and more grasps are also observed. \textbf{(c)} The agent can grasp from additional placements by changing the angle of the most distal joint, w2. The wrist orientation is copied from the final configuration of a trajectory that succeeded for a nearby placement (Section~\ref{sec:wrist}). The use of nearest neighbors applies best very close to existing successes, so most improvements can be observed in these areas. \textbf{(d)} Fine-tuning the grasp action (Section~\ref{subsection:fine_tuning}) provides the highest reliability of 72.5\%. The number of misses is also reduced to one, the lowest for any method considering features beyond those needed for reaching only. Most remaining nongrasps occur near the edges of the workspace where it is less natural to reach with the left hand, and where the PPS Graph is more sparse.} 
\label{fig:training_set_positions}
\vspace{1mm}
\hrule
\end{figure}

Finally, we observe that many failed grasps approach the top end of the target, despite being intended to move toward $c^t$. We hypothesize that the displacement comes from the difference between the perceived center, biased up and toward the camera from the observation of the front faces only, and the true center in 3D space. This suggests that altering the desired position from the perceived center, especially toward the true center, will result in a more precise grasp approach and additional successes. Following this suggestion of offsetting the intended position based on the length of the major axis of $t$, grid search or trial and error can discover the best performing offset over the training set. At this time, the highest success rate is achieved by decreasing the desired $v$ coordinate by a fourth of the major axis length (moving downward in the image), and decreasing $d$ and increasing $u$ by half of that amount (moving right and deeper). As seen in Figures~\ref{fig:GraspResult} and \ref{fig:training_set_positions}, using this offset and the other adjustments to the procedure yields a significant improvement in reliability to 72.5\%, with 29 grasps, six Palmar bumps that do not grasp, four bumps, and one miss over the training set of 40 object placements.

\subsection{Grasp Action Evaluation with New Target Positions}

Up to this point, all learning and evaluations starting with the baseline reach method with a random final node have used the same training set of object placements. Our agent has also assumed that a specific placement can be reset so that a reach or grasp attempt can be repeated with a new feature considered or value used. This allows a straightforward comparison of the results at each phase of learning, with the possibility of performance changes due to different random placements removed. However, the use of repeated placements also raises the question of how well the learned actions generalize, or whether the techniques are overfit to this training set.

We investigate this question for the agent's final version of the grasp action described in subsection \ref{subsection:fine_tuning}. Forty additional trials are conducted with the object placed in a distinct set of previously unseen locations, chosen from the same uniform random distribution of coordinates on the table surface. We conduct all trials of this experiment with the same block in an upright orientation, as before, and leave testing for generalization to different objects or orientations as future work.

The numerical results of this experiment are presented alongside the training set results in Figure~\ref{fig:GraspResult}. While grasps occurred much more often than by accident with the reach action, we see a decrease in performance on the test set to 47.5\% reliability, specifically 19 successful grasps, nine Palmar bumps that do not grasp, 11 bumps, and one miss. While a lower test performance is typical, this difference is highly significant (p = 0.0003), which suggests that not all of the grasping skill has generalized to the new set of poses. 

\begin{figure}
\hrule
\vspace{1mm}
\begin{center}
\includegraphics[width=\textwidth]{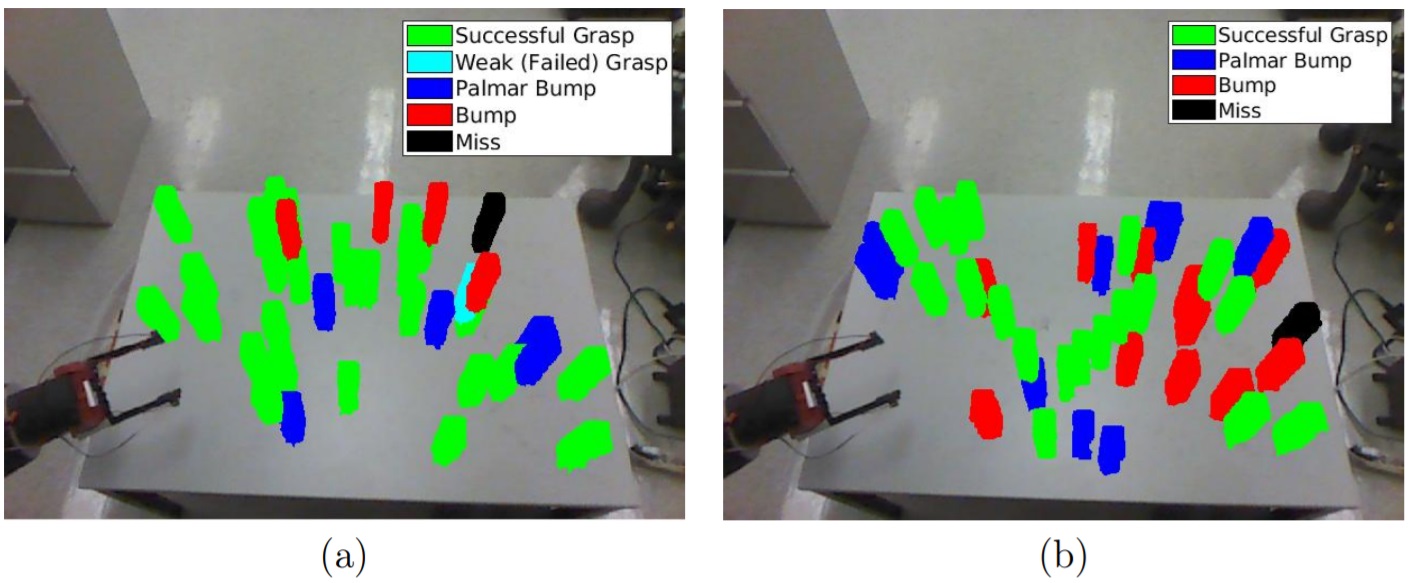}
\end{center}
\vspace{-4mm}
\caption{A spatial comparison of the results for the fine-tuned grasp action on the training and test sets. \textbf{(a)} Results for the training set, which the agent has had the ability to repeat trials with added features or to search over feature and fine-tuning values during learning. (Also shown in Figure~\ref{fig:training_set_positions}(d).) \textbf{(b)} Results for the test set of previously unseen target placements, using the same features and fine-tuning values developed on the training set. While still well above the performance of the reaching method baseline, the number of grasps is significantly reduced from the training to test sets. Some nongrasps occur in regions where the training set had no successes, and thus where the nearest neighbor orientation and fine-tuning values may have been incorrect. An alternative explanation is that some are in hard to reach regions where the PPS Graph of left hand configurations would be more sparse, providing fewer trajectory options and less informed estimates for adjusting the final motions.}
\label{fig:training_and_test_comparison}
\vspace{1mm}
\hrule
\end{figure}

Another possible explanation is that the test set of positions is more difficult to grasp the object from than the training set. The target placements for the training and test sets, colored according to the result of each trial, are compared in Figure~\ref{fig:training_and_test_comparison}. One measure for difficulty is the number of candidate final nodes available for a target placement, as this increases the likelihood of at least one candidate having $c^p_f$ very close to $c^t$ and $\vec{g}_f$ nearly perpendicular to $\vec{o}$. We found that the test set trials did have a lower mean number of candidate final nodes (41) than the training set trials (43), but the result was not statistically significant (p = 0.36), supporting the former explanation that the learned grasp action does not fully generalize to new positions. In future work, we will investigate which features are overfit to the training set and propose more general methods for improving grasp reliability. We believe that the adjustment to compensate for the perception of the target center should be investigated first, as the difference between the perceived and true centers may be dependent on the target location in the image, which is not yet considered, and the adjustment was based on the best results for the training set only.

\section{Conclusions}

We have demonstrated a computational model of an embodied learning agent, implemented on a physical Baxter robot, exploring its sensorimotor space without explicit guidance or feedback, constructing a representation of the robot’s peripersonal space (the PPS graph), including a mapping between the proprioceptive sensor and the visual sensor.  Its exploration learns the typical result of an action, but when an unusual result is observed, intrinsic motivation drives an effort to find conditions under which that result is reliable.

Using this approach, the agent first learns to accomplish reaching (and bumping an object to move it quasi-statically) with near 100\% reliability.  Importantly, during the early phases of learning, just like observations of human infants, the trajectories found are quite jerky, and the agent is able to reach an object even when the hand is not visible during the reach.  The final increase in action reliability comes from using the neighbors of a given node in the PPS graph to estimate the local Jacobian at that node, showing how to perturb the joint angles of the arm to accomplish a desired incremental motion of the image of the hand.  This fills in the gaps between the somewhat-sparsely-spaced nodes in the PPS graph, yielding highly accurate reaches.

Similarly, practice with reliable reaching leads to unusual accidental grasps (with assistance from the Palmar reflex), from which conditions on reliable grasping can be learned.  Grasping is substantially more complex than reaching, but the methods presented here approach 50\% reliability for intended grasps.  

In future research, we plan to pursue several opportunities for improvements to our model.  Our model of early visual perception is very simple.  Improvements to the grasp-learning model will require a model of how infant perceptual development improves the representation of object size and shape.  These are necessary to improve hand pre-shaping to suit the needs of grasping the specific object of interest.  Similarly, our simple model of controlled motion along edges in the PPS graph will be improved by a dynamical control model.  We expect this will improve our account of the jerkiness of early hand trajectories.  Using interpolation methods like the local Jacobian, we expect to be able to model how the reaching trajectory improves as children approach smooth, directed, adult reaches.

This preliminary model takes us a step toward understanding how an unguided agent can learn to control its own complex body, and objects in the space immediately around itself.

\bibliography{Juett_Kuipers}

\end{document}